\documentclass[
    a4paper,
    man,
    floatsintext
]{glossaPX2}

\usepackage[T1]{fontenc}
\usepackage[american]{babel}
\usepackage[style=apa,backend=biber,sorting=nyt,natbib=true]{biblatex}
\NewBibliographyString{unpublished}
\DefineBibliographyStrings{english}{unpublished = {Unpublished}}
\DefineBibliographyStrings{american}{unpublished = {Unpublished}}
\usepackage[font={footnotesize,it}]{caption}
\usepackage{csquotes}
\usepackage{booktabs}
\usepackage{tabularx}
\usepackage{linguex}
\usepackage{cgloss}

\usepackage{amsmath,amssymb,amsfonts,amsthm,mathrsfs}
\makeatletter
\tagsleft@false
\makeatother
\usepackage{graphicx}
\graphicspath{{figures/}}
\usepackage{ragged2e}
\usepackage{hyperref}
\usepackage{enumitem}
\usepackage[title]{appendix}
\usepackage{multirow}
\usepackage{makecell}
\usepackage{array}
\usepackage{adjustbox}
\usepackage{colortbl}
\usepackage{xcolor}
\usepackage{algorithm}
\usepackage{algpseudocode}
\usepackage{listings}
\usepackage{cleveref}
\usepackage{placeins}
\usepackage{float}
\usepackage{subcaption}
\usepackage{rotating}
\usepackage{url}
\usepackage{microtype}

\providecommand{\obs}{\mathbf{o}}
\providecommand{\act}{\mathbf{a}}
\providecommand{\robstate}{\mathbf{q}}
\providecommand{\demo}{\mathcal{D}}
\providecommand{\task}{\tau}
\providecommand{\ptask}{\mathbf{P}_{\task}}
\providecommand{\flow}{v_{\theta}}
\newcolumntype{L}{@{\extracolsep{\fill}}l}
\newcolumntype{C}{@{\extracolsep{\fill}}c}
\newcolumntype{R}{@{\extracolsep{\fill}}r}
\title[Task-Prototype Guided Flow Matching for Few-Shot Generalization in Vision-Language Robot Manipulation]{Task-Prototype Guided Flow Matching for Few-Shot Generalization in Vision-Language Robot Manipulation}
\author[Wang et al.]
{\spauthor{Yizhao Wang\\
  \institute{School of Computer Science, Henan Institute of Science and Technology}\\
  \small{cswyz@stu.hist.edu.cn, ORCID: 0009-0003-7056-7264}
  }
  \AND
\spauthor{Guantao Zhang\\
  \institute{School of Computer Science, Henan Institute of Science and Technology}\\
  \small{guantaozhang@hist.edu.cn}
  }
  \AND
\spauthor{Jingbo Wang\\
  \institute{School of Computer Science, Henan Institute of Science and Technology}\\
  \small{jingbowang@hist.edu.cn}
  }}

\begin{document}
\maketitle

\begin{abstract}
Vision-language robot manipulation policies can follow semantic instructions, but adapting them to a new procedure from only a few demonstrations remains difficult because language underspecifies contact timing, motion phases, corrective behavior, and execution style. This paper presents Task-Prototype Guided Flow Matching (TP-Flow), a few-shot manipulation framework that converts support demonstrations into structured task-prototype tokens and uses them to guide both the initial flow prior and the velocity field. TP-Flow employs symmetric cross-attention with learnable queries to extract phase-level prototypes, parameterizes a task-adaptive initial distribution, and injects prototype information through gated adaptive normalization. It is trained with an episodic support-query objective and prototype contrastive regularization, so few-shot adaptation is simulated during training while nuisance information is suppressed. On the LEROBOT-ARM-SO101 platform, TP-Flow achieves 66.8\%, 79.6\%, and 82.1\% success rates under 1-, 4-, and 6-shot settings, with a 75.5\% few-shot AUC. At 1-shot, it improves over CFM, Pooled-Demo CFM, and In-Context Flow by 29.8, 14.5, and 9.9 percentage points. It also improves held-out target-group generalization across novel-object transfer, goal recombination, long-horizon composition, and contact/correction tasks. TP-Flow maintains real-time execution with six online prototype tokens, 54.3 ms latency, 3.9 GB peak memory, and a 10 Hz control rate, while reducing the noisy-support success drop to 6.2\%. Theoretical diagnostics show that prototype distance aligns with action-distribution distance, the adaptive prior reduces transport cost, and gated modulation keeps measured trajectory deviations below the derived ODE bound. The code repository is omitted for anonymous review.
\end{abstract}


\section{Introduction}\label{sec:introduction}

General-purpose robot manipulation requires more than mapping a language instruction to a motor command. A deployed robot must infer how the task should be executed, including approach direction, grasp timing, contact handling, corrective motion, and release behavior. Vision-language-action (VLA) models have made substantial progress toward this goal by combining visual perception, language conditioning, and large-scale robot data. RT-1 demonstrates scalable transformer policies for real-world robot data \citep{brohan2022rt1}, RT-2 transfers web-scale vision-language knowledge to robotic control \citep{brohan2023rt2}, OpenVLA provides an open policy-modeling framework \citep{kim2024openvla}, and $\pi_0$ introduces flow-based action generation into VLA control \citep{black2024pi0}. However, semantic generalization alone does not guarantee that a robot can adapt to a new manipulation procedure from only a few demonstrations.

Few-shot adaptation is difficult because manipulation tasks contain execution details that are only weakly specified by language. The same instruction may require different grasp poses, clearance margins, contact phases, or recovery motions depending on object geometry, scene layout, and robot embodiment. Demonstrations provide these missing cues, but a small support set may also contain nuisance information such as lighting, background texture, accidental motion, or scene-specific object placement. Thus, the central question is not whether demonstrations are useful, but how a policy should represent them and how this representation should influence action generation. Existing benchmarks and software ecosystems, including CALVIN \citep{mees2022calvinbenchmarklanguageconditionedpolicy}, LIBERO \citep{liu2023liberobenchmarkingknowledgetransfer}, RoboMimic \citep{mandlekar2021matterslearningofflinehuman}, and LeRobot \citep{cadene2024lerobot}, further motivate reproducible few-shot robot-learning protocols as a complement to large-scale data collection. Recent control-engineering work on robotic force tracking and learning-assisted robot control also points to the importance of encoding execution structure rather than only task labels \citep{mu2025adaptiveimpedance,lu2025faulttolerantrl}.

Generative action policies provide a natural tool for this setting because many manipulation scenes admit multiple valid trajectories. Diffusion Policy shows that generative visuomotor policies can model multimodal action chunks \citep{chi2023diffusionpolicy}, while ACT highlights the stabilizing effect of temporal action chunking for fine-grained manipulation \citep{zhao2023act}. Flow matching offers a related formulation by learning a continuous velocity field that transports samples from a prior distribution to the expert action distribution \citep{lipman2023flowmatching}. Recent robot policies based on flow generation further suggest that flow-based action models are promising for fast continuous control \citep{zhang2024flowpolicyenablingfastrobust}. Yet ordinary conditional flow matching remains under-specified for few-shot cross-task adaptation.

Three issues are especially important. First, average pooling over demonstrations can discard the temporal structure that distinguishes approach, grasp, contact, correction, transport, and release. Second, using raw demonstrations as an in-context sequence increases online cost and may make the policy attend to irrelevant support details. Third, simple feature concatenation does not specify how strongly the support set should control the initial action prior or the velocity field. Overly strong conditioning may inject support noise into action generation, whereas overly weak conditioning may collapse to language-only control.

This paper studies a practical question: how can a robot reuse a shared manipulation prior while adapting to a new task from only a few demonstrations? We propose Task-Prototype Guided Flow Matching (TP-Flow), a few-shot robot manipulation framework that converts support demonstrations into structured task-prototype tokens. TP-Flow extracts phase-level prototypes with symmetric cross-attention, uses the prototypes to parameterize a task-adaptive initial flow prior, and injects prototype information into the velocity network through gated adaptive normalization. It is trained with an episodic support-query objective and prototype contrastive regularization, so few-shot adaptation is simulated during training while nuisance variation is suppressed.

The main contributions are summarized as follows:
\begin{itemize}
\item We propose TP-Flow, a few-shot flow-matching policy in which structured task-prototype tokens extracted from support demonstrations guide cross-task action generation.
\item We design a symmetric cross-attention prototype encoder that uses learnable queries to discover key observation-action phases from few-shot demonstrations, avoiding simple averaging of demo features.
\item We introduce prototype-guided flow generation, where prototypes parameterize a task-adaptive initial distribution and modulate the velocity field through gated adaptive normalization.
\item We train TP-Flow with an episodic support-query objective and a prototype contrastive loss, and provide the theoretical analysis and diagnostic experiments in the Supporting Material.
\end{itemize}

We evaluate TP-Flow on the LEROBOT-ARM-SO101 platform with few-shot adaptation, cross-task generalization, novel-object robustness, long-horizon transfer, inference efficiency, and module-level ablations.

\section{Related Work}\label{sec:related}

\subsection{Vision-language-action models for robot manipulation}

VLA models aim to map visual observations and language instructions directly to robot actions. RT-1 demonstrated that transformer-based robot policies can scale across a large collection of real-world tasks and robot data \citep{brohan2022rt1}. RT-2 further showed that vision-language knowledge learned from web-scale data can be transferred to robotic control \citep{brohan2023rt2}. OpenVLA made open-source VLA policy research more accessible \citep{kim2024openvla}, and Octo introduced a unified generalist policy interface for robot learning \citep{octomodelteam2024octoopensourcegeneralistrobot}. In parallel, $\pi_0$ introduced a VLA flow model for general robot control \citep{black2024pi0}, while TinyVLA investigated fast and data-efficient VLA modeling for manipulation \citep{tinyvla2025}. Recent Control Engineering Practice work on reinforcement-learning-assisted robot control reinforces the same trend toward learning executable robot behavior from data \citep{lu2025faulttolerantrl}. These models demonstrate the value of scalable multimodal policies, but most of them still rely on language, observations, or broad robot-data priors rather than an explicit few-shot execution prototype.

Several works show that language-conditioned manipulation is not only a question of high-level semantics, but also of representing actionable visual and spatial information. CLIPort combines semantic and spatial pathways for manipulation \citep{shridhar2021cliportpathwaysroboticmanipulation}; PerAct uses a multi-task transformer over voxelized observations \citep{shridhar2022perceiveractormultitasktransformerrobotic}; VIMA studies multimodal prompts for general robot manipulation \citep{jiang2023vimageneralrobotmanipulation}; and VoxPoser composes 3D value maps from language-model outputs \citep{huang2023voxposercomposable3dvalue}. Related recent control-engineering research makes the same point from a different angle: safe robot motion planning emphasizes executable spatial constraints \citep{nurbayeva2025safemotion}, mixed-reality teleoperation grounds commands in human-mobile manipulator interaction \citep{yeh2025teleoperation}, and robust integrated planning and control highlights the role of trajectory structure under disturbances \citep{huo2025integratedplanningcontrol}. These works show that task context must eventually be grounded in executable motion structure. TP-Flow follows this view, but focuses specifically on extracting execution structure from few-shot demonstrations rather than from language or prompt tokens alone.

Recent VLA and robot-learning systems expose several practical bottlenecks for embodied generalization. Open X-Embodiment emphasizes data diversity across embodiments \citep{embodimentcollaboration2025openxembodimentroboticlearning}, while DROID highlights the value and difficulty of in-the-wild robot data collection \citep{khazatsky2024droid}. RT-1 develops scalable real-world robot-data pretraining \citep{brohan2022rt1}; RT-2 connects web-scale vision-language knowledge to control \citep{brohan2023rt2}; OpenVLA focuses on open VLA policy modeling \citep{kim2024openvla}; Octo provides a generalist policy interface \citep{octomodelteam2024octoopensourcegeneralistrobot}; and TinyVLA studies a more compact VLA policy for efficient manipulation \citep{tinyvla2025}. VIMA shows that multimodal prompt structure can guide robot manipulation \citep{jiang2023vimageneralrobotmanipulation}, whereas $\pi_0$ uses continuous flow-based action generation in a VLA policy \citep{black2024pi0}. TP-Flow is complementary to these VLA developments. It does not require a new foundation backbone; rather, it introduces a structured few-shot conditioning mechanism that can be attached to flow-based action generation.

\subsection{Few-shot robot adaptation}

Few-shot robot adaptation seeks to acquire new manipulation skills from a small number of demonstrations. This setting is closely related to meta-learning, lifelong robot learning, representation transfer, and context-conditioned imitation. Meta-World provides a benchmark for multi-task and meta reinforcement learning \citep{yu2021metaworldbenchmarkevaluationmultitask}; RLBench supports many robot learning tasks in a unified environment \citep{james2019rlbenchrobotlearningbenchmark}; CALVIN focuses on language-conditioned long-horizon manipulation \citep{mees2022calvinbenchmarklanguageconditionedpolicy}; and LIBERO studies knowledge transfer and lifelong robot learning \citep{liu2023liberobenchmarkingknowledgetransfer}. In recent control-engineering settings, reinforcement-learning-assisted robot control and robust manipulator planning again underline that a few examples or repeated trials can carry a surprisingly rich task prior \citep{lu2025faulttolerantrl,huo2025integratedplanningcontrol}. These benchmarks collectively show that generalization should be evaluated not only by average success on seen tasks, but also by adaptation to new goals, scenes, and task compositions.

Representation learning is another route to few-shot adaptation. R3M demonstrates that reusable visual representations can improve downstream manipulation \citep{nair2022r3muniversalvisualrepresentation}, while RoboMimic studies what matters when learning from offline human demonstrations \citep{mandlekar2021matterslearningofflinehuman}. BC-Z explores zero-shot task generalization with robotic imitation learning \citep{jang2022bczzeroshottaskgeneralization}, and MT-Opt investigates large-scale multi-task reinforcement learning for continuous robotic control \citep{kalashnikov2021mtoptcontinuousmultitaskrobotic}. Single-demonstration imitation suggests that even one example can be informative when the model extracts stable execution cues \citep{li2025singledemonstration}. Long-horizon imitation further stresses that these cues must persist across extended temporal dependencies \citep{yao2025longhorizon}. SPECI studies a related continual-imitation setting through hierarchical skill prompts \citep{speci2026}. These methods show the value of shared representations and shared skill priors. However, they do not directly specify how a small support set should shape a generative action trajectory at test time.

Data scale and reproducibility are equally important. DROID provides large-scale in-the-wild robot manipulation data \citep{khazatsky2024droid}, while Open X-Embodiment aggregates heterogeneous robot datasets and RT-X models across embodiments \citep{embodimentcollaboration2025openxembodimentroboticlearning}. ALOHA demonstrates low-cost hardware for fine-grained bimanual manipulation \citep{zhao2023act}, and LeRobot provides an open-source library for end-to-end robot learning \citep{cadene2024lerobot}. TP-Flow is designed with this practical setting in mind: a shared policy is trained across tasks, while a small support set is used to compute a compact prototype representation for a target task.

In contrast to approaches that add demonstrations only as inference-time context, TP-Flow uses an episodic support-query objective. Each training episode samples a task, encodes support demonstrations into prototypes, and trains the action generator on query trajectories from the same task. This design explicitly matches the few-shot evaluation setting and reduces the mismatch between training and test-time adaptation.

\subsection{Generative policies and flow matching}

Generative policies are well suited to robot manipulation because many tasks admit multiple valid action trajectories. In a placing task, for example, several approach paths may be correct, while averaging those paths can produce an invalid collision trajectory. Diffusion Policy models action chunks through iterative denoising and has become a strong baseline for visuomotor policy learning \citep{chi2023diffusionpolicy}. FlowPolicy demonstrates that flow matching can support fast and robust manipulation policies \citep{zhang2024flowpolicyenablingfastrobust}. Affordance Flow Matching injects affordance information into flow-based robot manipulation \citep{zhang2025affordancebasedrobotmanipulationflow}. 3D FlowMatch Actor extends flow-based action generation to unified 3D single- and dual-arm policies \citep{gkanatsios20253dflowmatchactorunified}. ACT shows that temporal action chunking is important for fine-grained real-world manipulation \citep{zhao2023act}, and closed-loop grasp-generation work shows the complementary value of feedback during execution \citep{morrison2018closingloop}.

Flow matching provides an alternative to diffusion modeling by learning a velocity field that transports samples from a prior distribution to the expert action distribution \citep{lipman2023flowmatching}. In robotics, $\pi_0$ integrates flow-based action generation into a VLA policy \citep{black2024pi0}; FlowPolicy explores fast and robust 3D flow-based policies via consistency flow matching \citep{zhang2024flowpolicyenablingfastrobust}; Affordance Flow Matching combines affordance information with flow generation \citep{zhang2025affordancebasedrobotmanipulationflow}; and 3D FlowMatch Actor unifies 3D policies for single- and dual-arm manipulation \citep{gkanatsios20253dflowmatchactorunified}. These works support the relevance of flow-based policies for continuous control, but most of them treat task context as a standard condition.

TP-Flow differs from ordinary conditional flow policies in two ways. First, the task context is not a single pooled vector or a language embedding, but a set of structured prototype tokens extracted from few-shot observation-action demonstrations. Second, the prototypes control both the initial prior distribution and the internal velocity-field modulation. This distinction matters because conditional flow matching can otherwise become a generic ``condition plus velocity network'' formulation without a clear mechanism for how support demonstrations reshape the generated action trajectory.

\subsection{Task representation for cross-task generalization}

Task representations are crucial when transferring skills across manipulation tasks. A useful task representation should capture both semantic goals and execution-specific dynamics, including key frames, contact phases, correction cues, gripper timing, and action smoothness. ALOHA shows that action chunks help represent fine-grained temporal structure \citep{zhao2023act}. Diffusion Policy indicates that multimodal action distributions require generative sequence modeling rather than simple regression \citep{chi2023diffusionpolicy}. PerAct highlights the role of observation-action alignment in voxelized multi-task manipulation \citep{shridhar2022perceiveractormultitasktransformerrobotic}. CALVIN stresses long-horizon language-conditioned execution \citep{mees2022calvinbenchmarklanguageconditionedpolicy}, and RoboMimic shows that demonstration quality can strongly affect offline imitation performance \citep{mandlekar2021matterslearningofflinehuman}. Recent control-engineering work on force tracking, industrial-manipulator anomaly detection, and safe robot motion planning also suggests that contact geometry and predicted motion should sit inside the task representation itself \citep{mu2025adaptiveimpedance,alamir2025anomalydetection,nurbayeva2025safemotion}. This makes task representation especially important for few-shot adaptation.

Existing task-conditioning mechanisms can be roughly divided into language conditioning, global demonstration encoding, retrieval or in-context prompting, and learned task embeddings. Language conditioning is scalable but often under-specifies low-level execution. Global demonstration encoders are efficient but can lose temporal structure. In-context demonstration conditioning preserves more information, but inference cost grows with support length and the policy may attend to irrelevant support details. Learned task embeddings can improve adaptation, but their effect on a generative policy is often limited to feature concatenation. These design choices leave open a central question: how should few-shot demonstrations control the actual action-generation dynamics?

TP-Flow answers this question by representing each task as a set of structured prototype tokens extracted by symmetric cross-attention over observation-action demonstrations. The prototype set is intended to capture multiple manipulation phases rather than compressing the task into a single vector. A prototype contrastive objective encourages demonstrations of the same task to remain close despite scene variations, while different task dynamics remain separated. The resulting prototypes then guide both the flow prior and the gated velocity modulation, providing a more explicit bridge between few-shot task context and action generation. In this sense, TP-Flow is positioned between large VLA systems that rely on broad semantic priors and task-specific imitation methods that rely on dense demonstrations: it seeks to adapt a shared flow policy using a small but structured task context.

\section{Method}\label{sec:method}

\subsection{Problem setup and flow objective}\label{subsec:problem-flow}

We formulate few-shot vision-language robot manipulation as conditional generative action modeling. Let $\mathcal{T}=\{\task_i\}_{i=1}^{N}$ denote the source task set. At robot time step $t$, the policy receives an RGB observation $\obs_t$, a proprioceptive state $\robstate_t$, and a language instruction $\ell$, and predicts an action chunk $\act_{t:t+H}$ over horizon $H$. For a target task $\task$, the available support demonstrations are
\begin{equation}
\demo_{\task}^{S}
=
\{(\obs_{1:T}^{(k)},\robstate_{1:T}^{(k)},\act_{1:T}^{(k)},\ell^{(k)})\}_{k=1}^{K},
\quad
K\in\{1,2,4,6\}.
\label{eq:support-set}
\end{equation}
The support set is too small for reliable task-specific fine-tuning, but it contains execution cues that language alone may not specify, such as approach direction, contact timing, correction behavior, and release phase.

TP-Flow follows a support-query episodic protocol. Each training episode samples a task $\task$ and splits its demonstrations into a support set $\demo_{\task}^{S}$ and a disjoint query set $\demo_{\task}^{Q}$. A task encoder maps the support set into structured prototype tokens,
\begin{equation}
\ptask
=
g_{\phi}(\demo_{\task}^{S})
=
\{\mathbf{p}_{\task}^{1},\mathbf{p}_{\task}^{2},\ldots,\mathbf{p}_{\task}^{M}\}
\in\mathbb{R}^{M\times d},
\label{eq:abstract-prototype}
\end{equation}
where $M$ is the number of prototype tokens. Section~\ref{subsec:prototype} specifies how these tokens are constructed by symmetric cross-attention over observation-action demonstrations. The policy is therefore written as
\begin{equation}
\pi_{\theta}
(\act_{t:t+H}\mid \obs_t,\robstate_t,\ell,\demo_{\task}^{S})
\equiv
p_{\theta}(\act\mid \obs_t,\robstate_t,\ell,\ptask).
\label{eq:conditional-policy}
\end{equation}

Action generation is modeled with flow matching. Let $\mathbf{x}_1=\act^{\ast}$ be an expert action chunk sampled from the query set, and let $\mathbf{x}_0\sim p_0(\mathbf{x}\mid\ptask)$ be an initial action sample from a prototype-conditioned prior. For a flow time $s\in[0,1]$, TP-Flow uses the interpolation $\mathbf{x}_s=(1-s)\mathbf{x}_0+s\mathbf{x}_1$ with target velocity $\mathbf{u}_s=\mathbf{x}_1-\mathbf{x}_0$. The velocity field is trained by
\begin{equation}
\mathcal{L}_{\mathrm{FM}}
=
\mathbb{E}
\left[
\left\|
\flow(\mathbf{x}_s,s,\mathbf{h}_t,\ptask)
-
\mathbf{u}_s
\right\|_2^2
\right],
\label{eq:problem-fm-loss}
\end{equation}
where $\mathbf{h}_t=f_{\mathrm{obs}}(\obs_t,\robstate_t,\ell)$ is the current multimodal observation condition. At inference time, TP-Flow samples $\mathbf{x}_0$ from the prototype-guided prior and integrates $d\mathbf{x}_s/ds=\flow(\mathbf{x}_s,s,\mathbf{h}_t,\ptask)$ to obtain the action chunk.

This formulation exposes the gap addressed by TP-Flow. Ordinary conditional flow matching does not determine how few-shot demonstrations should be compressed, whether support-query adaptation should be simulated during training, or how strongly the support context should influence the prior and velocity field. TP-Flow resolves these choices through structured task prototypes, episodic training, a prototype-guided initial distribution, and gated prototype modulation.

\subsection{Overview}\label{subsec:overview}

\begin{figure}[tp]
\centering
\includegraphics[width=\textwidth]{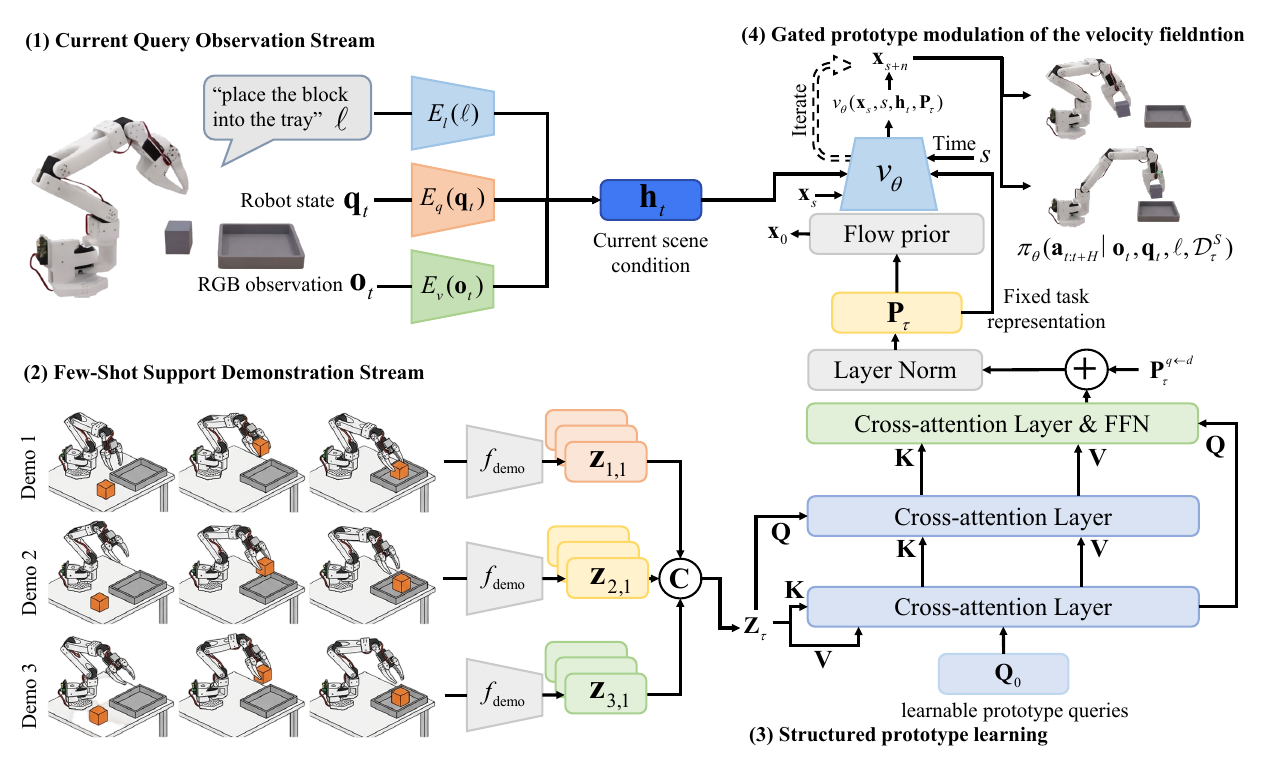}
\caption{Overview of TP-Flow. Few-shot support demonstrations are encoded into structured task-prototype tokens. The prototypes parameterize a task-adaptive flow prior and modulate the velocity field through gated adaptive normalization.}
\label{fig:framework}
\end{figure}

Figure~\ref{fig:framework} illustrates the overall TP-Flow pipeline. Given a current observation $(\obs_t,\robstate_t)$, a language instruction $\ell$, and a few-shot support set $\demo_{\task}^{S}$, the model generates a future action chunk $\act_{t:t+H}$. The method contains four coupled components. First, a multimodal encoder converts the current robot state and each support demonstration frame into observation-action tokens. Second, a symmetric cross-attention encoder compresses the support set into a fixed number of structured task-prototype tokens $\ptask$. Third, the prototype tokens parameterize a task-adaptive flow prior from which the initial action sample $\mathbf{x}_0$ is drawn. Fourth, the same prototype tokens modulate the velocity field through gated adaptive normalization.

This design differs from ordinary conditional flow matching in three ways. The support set is not represented by a single averaged feature; instead, TP-Flow keeps $M$ prototype tokens so that different manipulation phases can be represented separately. The prototype is not used only as an additional condition; it changes both the initial prior distribution and the internal velocity-field computation. Finally, the few-shot setting is used during training through support-query episodes rather than introduced only as an inference-time prompt.

\subsection{Multimodal observation encoding}\label{subsec:obs-encoder}

\begin{figure}[htbp]
\centering
\includegraphics[width=\linewidth]{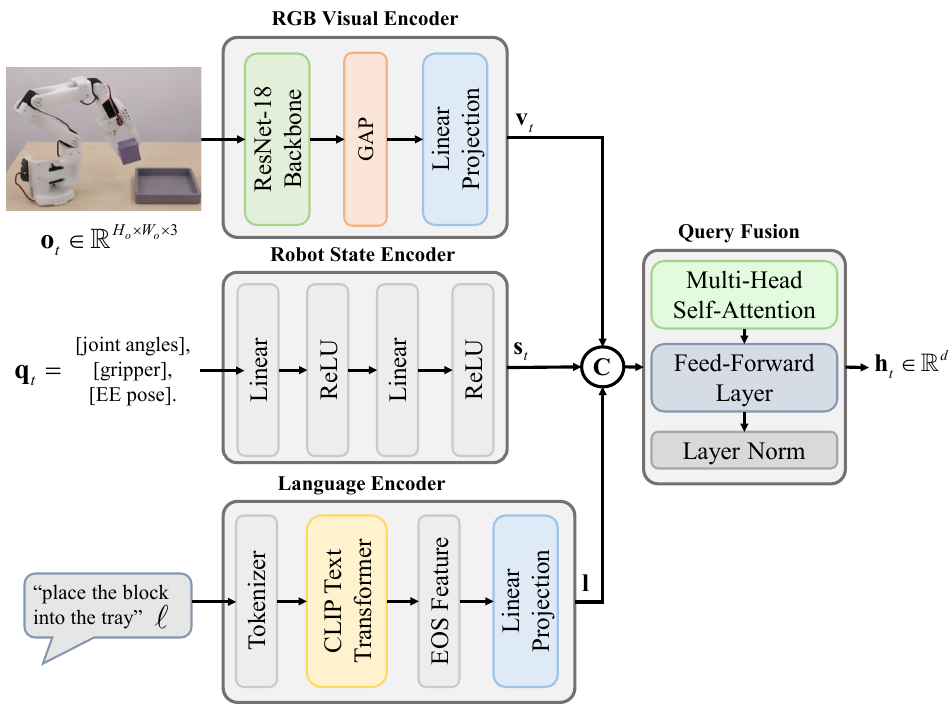}
\caption{Internal structure of the shared multimodal observation encoder $f_{\mathrm{obs}}$. The RGB branch, robot-state branch, and language branch produce $\mathbf{v}_t$, $\mathbf{s}_t$, and $\mathbf{l}$, which are fused by a query-fusion block to obtain the current condition $\mathbf{h}_t$. The same encoder weights are reused for support frames to compute $\mathbf{h}_{k,j}^{S}$; action and temporal information are then added by the support-token construction in Eqs.~\ref{eq:support-action-token}--\ref{eq:demo-token}.}
\label{fig:multimodal-encoder}
\end{figure}

TP-Flow uses a shared multimodal observation encoder for both the current query state and the visual-proprioceptive part of each support demonstration. Figure~\ref{fig:multimodal-encoder} details this shared encoder. It maps RGB observation, robot state, and language into a unified condition token through visual, proprioceptive, and text branches followed by query fusion. For the current query state, we write
\begin{equation}
\mathbf{h}_t
=
f_{\mathrm{obs}}(\obs_t,\robstate_t,\ell)
\in \mathbb{R}^{d},
\label{eq:query-encoding}
\end{equation}
where $\obs_t$ is the RGB observation at time $t$, $\robstate_t$ is the proprioceptive state, $\ell$ is the language instruction, and $d$ is the hidden dimension. The vector $\mathbf{h}_t$ represents what the robot currently observes and what semantic goal it should pursue. It does not by itself contain the few-shot execution pattern; that information is provided by $\ptask$.

For the $k$-th support demonstration, the same observation encoder is reused at the $j$-th step:
\begin{equation}
\mathbf{h}_{k,j}^{S}
=
f_{\mathrm{obs}}
(\obs_{j}^{(k)},\robstate_{j}^{(k)},\ell_{\task})
\in \mathbb{R}^{d},
\label{eq:support-observation-token}
\end{equation}
where $\ell_{\task}$ denotes the task instruction associated with the support set. The action $\act_{j}^{(k)}$ is then encoded by an action MLP,
\begin{equation}
\mathbf{r}_{k,j}
=
E_a(\act_{j}^{(k)})
\in \mathbb{R}^{d}.
\label{eq:support-action-token}
\end{equation}
The support observation-action token is constructed as
\begin{equation}
\mathbf{z}_{k,j}
=
F_{\mathrm{demo}}
\left(
\mathbf{h}_{k,j}^{S},
\mathbf{r}_{k,j},
\mathbf{e}_{j},
\mathbf{e}_{k}^{\mathrm{demo}}
\right)
\in \mathbb{R}^{d},
\label{eq:demo-token}
\end{equation}
where $\mathbf{e}_{j}$ is a temporal positional embedding and $\mathbf{e}_{k}^{\mathrm{demo}}$ identifies the support demonstration. This partial weight sharing aligns the query state and the support frames in the same multimodal representation space, while the support-specific action-temporal fusion module models gripper timing and local motion tendency.

All support tokens are stacked into
\begin{equation}
\mathbf{Z}_{\task}
=
\{\mathbf{z}_{k,j}\}_{k=1,j=1}^{K,T_k}
\in
\mathbb{R}^{L\times d},
\label{eq:support-token-set}
\end{equation}
where $L=\sum_{k=1}^{K}T_k'$ is the number of support tokens after optional temporal subsampling, and $T_k'$ is the retained length of the $k$-th demonstration. This token set is the input to the prototype encoder.

\subsection{Structured prototype learning with symmetric cross-attention}\label{subsec:prototype}

The goal of the prototype encoder is to map the variable-length support token set $\mathbf{Z}_{\task}$ to a fixed-size task representation $\ptask\in\mathbb{R}^{M\times d}$. Here, $M$ is the number of prototype tokens. In practice, $M$ is much smaller than $L$, so the prototype encoder compresses the few-shot demonstrations while preserving phase-level execution cues.

We define a standard cross-attention operator as
\begin{equation}
\mathrm{Attn}(\mathbf{Q},\mathbf{K},\mathbf{V})
=
\mathrm{softmax}
\left(
\frac{\mathbf{Q}\mathbf{K}^{\top}}{\sqrt{d}}
\right)
\mathbf{V},
\label{eq:attn}
\end{equation}
where $\mathbf{Q}$ is the query matrix, $\mathbf{K}$ is the key matrix, $\mathbf{V}$ is the value matrix, and $d$ is the attention dimension. The softmax term computes how strongly each query token attends to each key token, and the final product returns a weighted combination of value tokens.

TP-Flow starts from $M$ learnable prototype queries:
\begin{equation}
\mathbf{Q}_{0}
=
\{\mathbf{q}_{1},\mathbf{q}_{2},\ldots,\mathbf{q}_{M}\}
\in \mathbb{R}^{M\times d}.
\label{eq:learnable-query}
\end{equation}
These queries are not tied to fixed semantic labels. During training, they learn to specialize to reusable manipulation phases, such as approach, grasp, contact, correction, transport, and release.

The first direction of attention lets the prototype queries retrieve information from support demonstrations:
\begin{equation}
\mathbf{P}_{\task}^{q\leftarrow d}
=
\mathrm{Attn}(\mathbf{Q}_{0},\mathbf{Z}_{\task},\mathbf{Z}_{\task})
\in \mathbb{R}^{M\times d}.
\label{eq:q-to-demo}
\end{equation}
In Eq.~\ref{eq:q-to-demo}, each learnable query acts as a soft detector over all support tokens. For example, one query may place high attention on frames where the gripper closes, while another may attend to corrective motions after contact.

The second direction lets support tokens attend back to the query-derived prototype tokens:
\begin{equation}
\mathbf{Z}_{\task}^{d\leftarrow q}
=
\mathrm{Attn}(\mathbf{Z}_{\task},\mathbf{P}_{\task}^{q\leftarrow d},\mathbf{P}_{\task}^{q\leftarrow d})
\in \mathbb{R}^{L\times d}.
\label{eq:demo-to-q}
\end{equation}
This reverse attention provides a consistency check: if a prototype token is useful, it should also explain or reconstruct task-relevant information in the support tokens. Background, lighting, or accidental visual details are less likely to remain stable under this bidirectional interaction.

The final structured prototype set is computed by residual fusion and feed-forward refinement:
\begin{equation}
\ptask
=
\mathrm{LN}
\left(
\mathbf{P}_{\task}^{q\leftarrow d}
+
\mathrm{FFN}
\left[
\mathrm{Attn}
(\mathbf{P}_{\task}^{q\leftarrow d},
\mathbf{Z}_{\task}^{d\leftarrow q},
\mathbf{Z}_{\task}^{d\leftarrow q})
\right]
\right)
\in \mathbb{R}^{M\times d}.
\label{eq:prototype}
\end{equation}
Here, $\mathrm{LN}$ denotes layer normalization and $\mathrm{FFN}$ denotes a feed-forward network. Eq.~\ref{eq:prototype} produces a fixed-size prototype set regardless of the number or length of support demonstrations. This is the key distinction from average-pooling support features: TP-Flow keeps multiple prototype tokens so that different execution phases can influence action generation differently.

\subsection{Prototype-guided flow prior}\label{subsec:prior}

Standard flow matching often samples the initial point from an isotropic Gaussian distribution, $\mathbf{x}_0\sim\mathcal{N}(0,I)$. In robot manipulation, this assumption can be unnecessarily loose. A high-precision insertion task may require a narrow initial action distribution, while a placing task may permit broader diversity. TP-Flow therefore lets the support-derived prototype guide the initial action prior.

We first pool the prototype set into a global task summary:
\begin{equation}
\bar{\mathbf{p}}_{\task}
=
\mathrm{Pool}(\ptask)
\in \mathbb{R}^{d},
\label{eq:prototype-pool}
\end{equation}
where $\mathrm{Pool}(\cdot)$ can be mean pooling or attention pooling over the $M$ prototype tokens. The vector $\bar{\mathbf{p}}_{\task}$ summarizes the overall task execution style.

The prior network predicts a mean and a raw log-variance vector:
\begin{equation}
\boldsymbol{\mu}_{\task},\hat{\boldsymbol{\sigma}}_{\task}
=
f_{\mathrm{prior}}(\bar{\mathbf{p}}_{\task}),
\label{eq:prior-params}
\end{equation}
where $\boldsymbol{\mu}_{\task}\in\mathbb{R}^{D}$ and $\hat{\boldsymbol{\sigma}}_{\task}\in\mathbb{R}^{D}$ have the same dimension as the flattened action chunk, with $D=H d_a$ and $d_a$ denoting the single-step action dimension. The mean $\boldsymbol{\mu}_{\task}$ shifts the starting distribution toward a task-specific action manifold.

To avoid variance collapse or overly diffuse sampling, the standard deviation is bounded:
\begin{equation}
\boldsymbol{\sigma}_{\task}
=
\sigma_{\min}
+
(\sigma_{\max}-\sigma_{\min})
\cdot
\mathrm{sigmoid}(\hat{\boldsymbol{\sigma}}_{\task}),
\label{eq:bounded-sigma}
\end{equation}
where $\sigma_{\min}$ and $\sigma_{\max}$ are scalar hyperparameters. Eq.~\ref{eq:bounded-sigma} prevents the prior from becoming deterministic and prevents the prototype from producing extremely noisy initial samples.

The initial flow sample is then drawn by the reparameterization:
\begin{equation}
\mathbf{x}_0
=
\boldsymbol{\mu}_{\task}
+
\boldsymbol{\sigma}_{\task}\odot\boldsymbol{\epsilon},
\quad
\boldsymbol{\epsilon}\sim\mathcal{N}(0,I).
\label{eq:adaptive-prior}
\end{equation}
Here, $\odot$ denotes element-wise multiplication. The prototype-guided prior is important because it changes the starting point of the continuous action-generation process. Thus, the support demonstrations affect the whole flow trajectory, not only the final velocity prediction.

\subsection{Gated prototype modulation of the velocity field}\label{subsec:flow}

Let $\mathbf{x}_1=\act^{\ast}$ denote an expert action chunk from the query set, and let $\mathbf{x}_0$ denote a sample from Eq.~\ref{eq:adaptive-prior}. Following the flow-matching formulation in Section~\ref{subsec:problem-flow}, we construct an interpolation:
\begin{equation}
\mathbf{x}_s
=
(1-s)\mathbf{x}_0+s\mathbf{x}_1,
\quad
s\in[0,1],
\label{eq:method-interpolation}
\end{equation}
with target velocity
\begin{equation}
\mathbf{u}_s
=
\mathbf{x}_1-\mathbf{x}_0.
\label{eq:method-target-velocity}
\end{equation}
The scalar $s$ is the flow time, not the robot time step. When $s=0$, the sample is drawn from the prototype-guided prior; when $s=1$, the sample reaches the expert action chunk. The velocity network learns how to move intermediate samples toward expert-like actions.

The velocity field is conditioned on three information sources:
\begin{equation}
\flow
(\mathbf{x}_s,s,\mathbf{h}_t,\ptask)
\in \mathbb{R}^{D}.
\label{eq:velocity-field}
\end{equation}
Here, $\mathbf{x}_s$ is the current action sample, $s$ tells the network where it is along the flow path, $\mathbf{h}_t$ describes the current observation and instruction, and $\ptask$ provides the few-shot execution context.

\begin{figure}[htbp]
\centering
\includegraphics[width=0.70\linewidth]{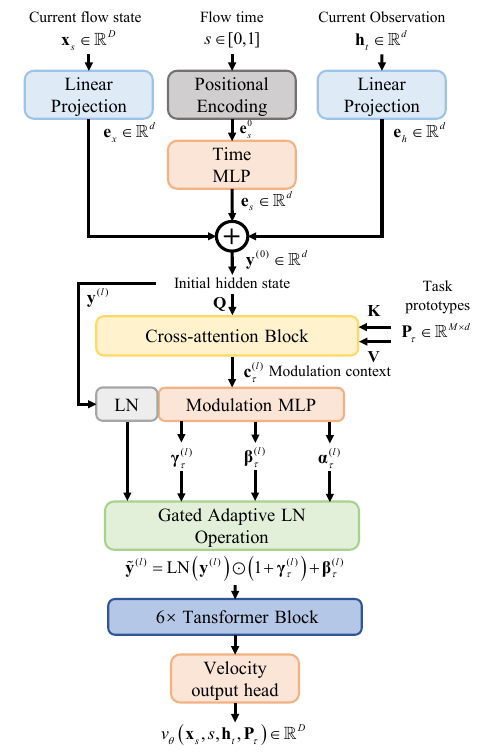}
\caption{Internal structure of the gated prototype-modulated velocity network. The flow state $\mathbf{x}_s$, flow time $s$, and observation condition $\mathbf{h}_t$ are embedded into $\mathbf{y}^{(0)}$. Each velocity layer queries the task prototypes $\ptask$ to generate scale, shift, and gate parameters for gated adaptive normalization, and the output head predicts $v_{\theta}(\mathbf{x}_s,s,\mathbf{h}_t,\ptask)$.}
\label{fig:flow-generation}
\end{figure}

Instead of concatenating $\ptask$ to every hidden feature, TP-Flow uses the gated prototype-modulated velocity network in Fig.~\ref{fig:flow-generation}. The network first embeds the current flow state, flow time, and current observation condition into a common hidden space:
\begin{equation}
\begin{aligned}
\mathbf{e}_{\mathrm{state}} &=
\mathbf{W}_{x}\mathbf{x}_{s}+\mathbf{b}_{x},\\
\mathbf{e}_{\mathrm{time}} &=
\operatorname{MLP}_{s}(\operatorname{PE}(s)),\\
\mathbf{e}_{\mathrm{obs}} &=
\mathbf{W}_{h}\mathbf{h}_{t}+\mathbf{b}_{h},\\
\mathbf{y}^{(0)} &=
\mathbf{e}_{\mathrm{state}}+\mathbf{e}_{\mathrm{time}}+\mathbf{e}_{\mathrm{obs}}
\in\mathbb{R}^{d}.
\end{aligned}
\label{eq:velocity-input-embedding}
\end{equation}
Here, $\mathbf{e}_{\mathrm{state}}$ embeds the interpolated action state $\mathbf{x}_s$, $\mathbf{e}_{\mathrm{time}}$ embeds the flow time $s$, and $\mathbf{e}_{\mathrm{obs}}$ embeds the current observation-language condition. The three symbols are kept distinct so that the initial hidden state contains state, time, and observation information simultaneously.
For the $l$-th velocity block, the hidden state queries the prototype set through cross-attention:
\begin{equation}
\mathbf{c}_{\task}^{(l)}
=
\operatorname{Attn}
\left(
\mathbf{y}^{(l)},\ptask,\ptask
\right),
\label{eq:velocity-prototype-context}
\end{equation}
where the query is $\mathbf{y}^{(l)}$ and the keys and values are the prototype tokens $\ptask$. The modulation network then predicts layer-wise scale, shift, and gate vectors:
\begin{equation}
\boldsymbol{\gamma}_{\task}^{(l)},
\boldsymbol{\beta}_{\task}^{(l)},
\boldsymbol{\alpha}_{\task}^{(l)}
=
\operatorname{MLP}_{\mathrm{gate}}^{(l)}
\left(
\mathbf{c}_{\task}^{(l)}
\right),
\label{eq:mod-params}
\end{equation}
where $\boldsymbol{\gamma}_{\task}^{(l)}$ is a scale vector, $\boldsymbol{\beta}_{\task}^{(l)}$ is a shift vector, and $\boldsymbol{\alpha}_{\task}^{(l)}\in[0,1]^d$ is a gate vector. In practice, the gate is obtained with a sigmoid activation so that noisy support prototypes can be softly attenuated.

Adaptive normalization is applied as:
\begin{equation}
\widetilde{\mathbf{y}}^{(l)}
=
\operatorname{LN}
\left(
\mathbf{y}^{(l)}
\right)
\odot
\left(
1+\boldsymbol{\gamma}_{\task}^{(l)}
\right)
+
\boldsymbol{\beta}_{\task}^{(l)}.
\label{eq:adaln}
\end{equation}
The normalized feature $\operatorname{LN}(\mathbf{y}^{(l)})$ removes feature-scale variation, while $\boldsymbol{\gamma}_{\task}^{(l)}$ and $\boldsymbol{\beta}_{\task}^{(l)}$ restore a task-specific scale and bias. This lets the prototype alter the internal computation without forcing raw support tokens into the velocity network.

The modulated feature is injected through a gated residual path:
\begin{equation}
\begin{aligned}
\mathbf{m}^{(l)}
&=
F_{\theta}^{(l)}
\left(
\widetilde{\mathbf{y}}^{(l)}
\right),\\
\mathbf{y}^{(l+1)}
&=
\mathbf{y}^{(l)}
+
\boldsymbol{\alpha}_{\task}^{(l)}
\odot
\mathbf{m}^{(l)}.
\end{aligned}
\label{eq:gated-modulation}
\end{equation}
The function $F_{\theta}^{(l)}$ denotes a local transformation block, such as a feed-forward network, transformer block, or residual block. The gate $\boldsymbol{\alpha}_{\task}^{(l)}$ controls how much prototype-modulated information enters the hidden state. If the support demonstrations are noisy or visually inconsistent, the model can reduce the injection strength. If the task prototype is informative, the model can use a stronger modulation. After stacking $L_v$ velocity blocks, the output head predicts
\begin{equation}
\flow(\mathbf{x}_s,s,\mathbf{h}_t,\ptask)
=
\operatorname{Head}
\left(
\mathbf{y}^{(L_v)}
\right)
\in\mathbb{R}^{D}.
\label{eq:velocity-head}
\end{equation}
This layer-wise mechanism is more flexible than hard concatenation and more explicit than language-only conditioning.

The flow matching loss is
\begin{equation}
\mathcal{L}_{\mathrm{FM}}
=
\mathbb{E}_{\task,\demo_{\task}^{S},\demo_{\task}^{Q},s,\mathbf{x}_0,\mathbf{x}_1}
\left[
\left\|
\flow(\mathbf{x}_s,s,\mathbf{h}_t,\ptask)
-
\mathbf{u}_s
\right\|_2^2
\right].
\label{eq:fm-loss}
\end{equation}
The expectation is taken over sampled tasks, support-query episodes, interpolation time, initial prior samples, and expert action chunks. The loss trains the velocity field to match the ground-truth transport direction from $\mathbf{x}_0$ to $\mathbf{x}_1$.

During inference, an action chunk is generated by numerically integrating the learned velocity field:
\begin{equation}
\frac{d\mathbf{x}_s}{ds}
=
\flow(\mathbf{x}_s,s,\mathbf{h}_t,\ptask),
\quad
\mathbf{x}_{0}\sim\mathcal{N}(\boldsymbol{\mu}_{\task},\mathrm{diag}(\boldsymbol{\sigma}_{\task}^{2})).
\label{eq:inference-ode}
\end{equation}
The final sample $\mathbf{x}_{1}$ is reshaped into the action chunk $\act_{t:t+H}$.

\subsection{Episodic training and prototype contrastive regularization}\label{subsec:regularization}

TP-Flow is trained episodically. In each iteration, a task is sampled and its demonstrations are divided into support and query sets. The support set produces $\ptask$, while the query set provides observation-action pairs for the flow matching objective. This simulates the few-shot adaptation setting during training rather than adding demonstrations only at test time.

Few-shot prototypes can overfit to background, lighting, or accidental object poses. To reduce this risk, we sample two different support views from the same task, $\demo_{\task}^{S,a}$ and $\demo_{\task}^{S,b}$, and encode prototypes $\mathbf{P}_{\task}^{a}$ and $\mathbf{P}_{\task}^{b}$. We then pool them into
\begin{equation}
\bar{\mathbf{p}}_{\task}^{a}
=
\mathrm{Pool}(\mathbf{P}_{\task}^{a}),
\qquad
\bar{\mathbf{p}}_{\task}^{b}
=
\mathrm{Pool}(\mathbf{P}_{\task}^{b}).
\label{eq:contrastive-pool}
\end{equation}
The prototype contrastive loss for task $i$ is
\begin{equation}
\mathcal{L}_{\mathrm{proto}}
=
-\log
\frac{
\exp(\mathrm{sim}(\bar{\mathbf{p}}_{i}^{a},\bar{\mathbf{p}}_{i}^{b})/\gamma)
}{
\sum_{j}
\exp(\mathrm{sim}(\bar{\mathbf{p}}_{i}^{a},\bar{\mathbf{p}}_{j}^{b})/\gamma)
},
\label{eq:proto-loss}
\end{equation}
where $\gamma$ is a temperature parameter and $\mathrm{sim}(\cdot,\cdot)$ is cosine similarity. The numerator pulls two support encodings of the same task together, while the denominator pushes prototypes from different tasks apart. This encourages $\ptask$ to encode task-level execution structure rather than incidental visual details.

We also regularize the adaptive prior to prevent degenerate variance collapse:
\begin{equation}
\mathcal{L}_{\mathrm{prior}}
=
D_{\mathrm{KL}}
\left(
\mathcal{N}(\boldsymbol{\mu}_{\task},\mathrm{diag}(\boldsymbol{\sigma}_{\task}^{2}))
\;\|\;
\mathcal{N}(0,I)
\right).
\label{eq:prior-loss}
\end{equation}
This KL term keeps the prototype-guided prior close to a well-behaved Gaussian unless the support demonstrations provide evidence for a strong task-specific shift.

Finally, we add an auxiliary velocity smoothness regularizer:
\begin{equation}
\mathcal{L}_{\mathrm{smooth}}
=
\mathbb{E}_{s,\Delta}
\left[
\left\|
\flow(\mathbf{x}_s,s+\Delta,c_{\task,t})
-
\flow(\mathbf{x}_s,s,c_{\task,t})
\right\|_2^2
\right],
\label{eq:smooth-loss}
\end{equation}
where $c_{\task,t}=(\mathbf{h}_t,\ptask)$ and $\Delta$ is a small flow-time perturbation. This term discourages abrupt velocity changes along the generation trajectory.

The local prototype-consistency regularizer $\mathcal{L}_{\mathrm{loc}}$ is computed over task pairs in each episodic batch, as detailed in the Supporting Material. It aligns prototype neighborhoods with empirical action-distribution neighborhoods and prevents the prototype encoder from separating support sets whose query action statistics are locally similar.

The full training objective is
\begin{equation}
\mathcal{L}
=
\mathcal{L}_{\mathrm{FM}}
+
\lambda_{\mathrm{proto}}\mathcal{L}_{\mathrm{proto}}
+
\lambda_{\mathrm{prior}}\mathcal{L}_{\mathrm{prior}}
+
\lambda_{\mathrm{smooth}}\mathcal{L}_{\mathrm{smooth}}
+
\lambda_{\mathrm{loc}}\mathcal{L}_{\mathrm{loc}}.
\label{eq:full-loss}
\end{equation}
The coefficients $\lambda_{\mathrm{proto}}$, $\lambda_{\mathrm{prior}}$, $\lambda_{\mathrm{smooth}}$, and $\lambda_{\mathrm{loc}}$ balance prototype invariance, prior stability, velocity smoothness, and local prototype consistency against the main flow-matching objective.

\subsection{Training and inference}\label{subsec:training}

Algorithm~\ref{alg:training} summarizes the episodic training procedure of TP-Flow, including support-query task sampling, prototype extraction, adaptive-prior estimation, flow-matching supervision, prototype contrastive regularization, local prototype consistency, and velocity-smoothness regularization.

\begin{algorithm}[t]
\caption{Episodic training of TP-Flow}
\label{alg:training}
\begin{algorithmic}[1]
\Require Demonstration tasks $\{\demo_{\task_i}\}_{i=1}^{N}$; $K\in\{1,2,4,6\}$; $M$ prototypes; weights $\lambda_{\mathrm{proto}},\lambda_{\mathrm{prior}},\lambda_{\mathrm{smooth}},\lambda_{\mathrm{loc}}$
\Require Parameters $\Theta=\{\phi,\psi,\omega,\theta\}$ of $f_{\mathrm{obs}}$, $f_{\psi}$, $f_{\mathrm{prior}}$, and $\flow$
\For{each training iteration}
    \State Sample episodic batch $\mathcal{B}\sim p(\task)$ and split each task into $(\demo_{\task_i}^{S,a},\demo_{\task_i}^{S,b},\demo_{\task_i}^{Q})$
    \State Encode support views as $\mathbf{Z}_{\task_i}^{r}=f_{\mathrm{demo}}(\demo_{\task_i}^{S,r})$, $\mathbf{P}_{\task_i}^{r}=f_{\psi}(\mathbf{Z}_{\task_i}^{r})$, and $\bar{\mathbf{p}}_{\task_i}^{r}=\operatorname{Pool}(\mathbf{P}_{\task_i}^{r})$ for $r\in\{a,b\}$
    \State Estimate prototype-guided prior $(\boldsymbol{\mu}_{\task_i},\hat{\boldsymbol{\sigma}}_{\task_i})=f_{\mathrm{prior}}(\bar{\mathbf{p}}_{\task_i}^{a})$ and $\boldsymbol{\sigma}_{\task_i}=\operatorname{Bound}(\hat{\boldsymbol{\sigma}}_{\task_i})$
    \State For query chunks, sample $\boldsymbol{\epsilon}\sim\mathcal{N}(0,I)$ and $s\sim\mathcal{U}(0,1)$:
    \State Set $\mathbf{h}_t=f_{\mathrm{obs}}(\obs_t,\robstate_t,\ell)$, $\mathbf{x}_1=\act_{t:t+H}^{\ast}$, $\mathbf{x}_0=\boldsymbol{\mu}_{\task_i}+\boldsymbol{\sigma}_{\task_i}\odot\boldsymbol{\epsilon}$, and $\mathbf{x}_s=(1-s)\mathbf{x}_0+s\mathbf{x}_1$
    \State Predict velocity with $\mathbf{u}_s=\mathbf{x}_1-\mathbf{x}_0$ and $\hat{\mathbf{u}}_s=\flow(\mathbf{x}_s,s,\mathbf{h}_t,\mathbf{P}_{\task_i}^{a})$
    \State Compute task-pair distances $d_P$ and $d_A$ in the episodic batch, and evaluate $\mathcal{L}_{\mathrm{loc}}$
    \State Compute $\mathcal{L}_{\mathrm{epi}}=\mathbb{E}\|\hat{\mathbf{u}}_s-\mathbf{u}_s\|_2^2+\lambda_{\mathrm{proto}}\mathcal{L}_{\mathrm{proto}}+\lambda_{\mathrm{prior}}\mathcal{L}_{\mathrm{prior}}+\lambda_{\mathrm{smooth}}\mathcal{L}_{\mathrm{smooth}}+\lambda_{\mathrm{loc}}\mathcal{L}_{\mathrm{loc}}$
    \State Update $\Theta\leftarrow\Theta-\eta\nabla_{\Theta}\mathcal{L}_{\mathrm{epi}}$ with AdamW
\EndFor
\end{algorithmic}
\end{algorithm}

At test time, the robot receives $K$ demonstrations of a new task, computes $\ptask$ once, and reuses the prototype for all subsequent query observations. For each control step, TP-Flow samples $\mathbf{x}_0$ from the prototype-guided prior and integrates Eq.~\ref{eq:inference-ode}. Only the first action or a short prefix of the generated chunk is executed before replanning. Since the raw demonstrations are compressed into $M$ prototype tokens, inference cost is controlled by $M$ rather than by the full support trajectory length.

\noindent\textbf{Theoretical analysis and diagnostic experiments.}
The detailed theoretical analysis and its corresponding diagnostic experiments are provided in the Supporting Material. In brief, the analysis studies local prototype consistency, the variance-reduction effect of the prototype-adaptive prior, and the stability benefit of gated prototype modulation under noisy support demonstrations. The supporting experiments report prototype/action-distance correlation, adaptive-prior transport-cost diagnostics, and gated ODE perturbation tests on LEROBOT-ARM-SO101.

\section{Experiments}\label{sec:experiments}

\subsection{Experimental protocol and task design}\label{subsec:setup}

\begin{figure}[htbp]
\centering
\includegraphics[width=0.58\linewidth]{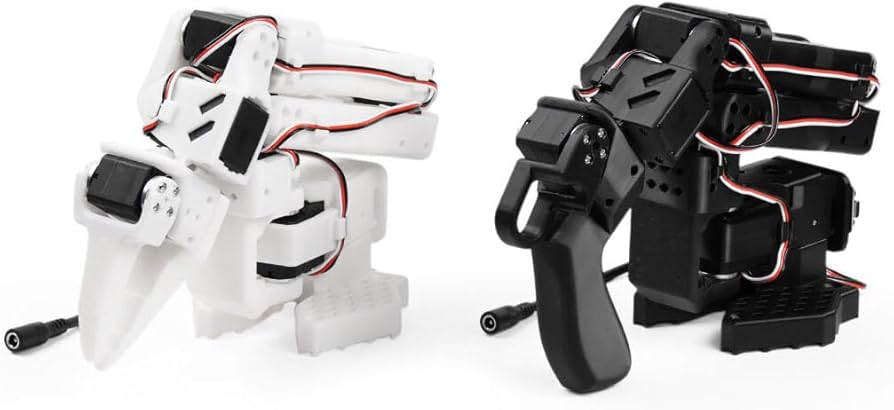}
\caption{LEROBOT-ARM-SO101 robotic platform used in all experiments. The platform provides the visual-proprioceptive observation stream and gripper-based manipulation interface used by TP-Flow.}
\label{fig:lerobot-arm-so101}
\end{figure}

We evaluate TP-Flow on a single experimental platform, LEROBOT-ARM-SO101, shown in Fig.~\ref{fig:lerobot-arm-so101}. No additional benchmark, robot embodiment, or custom tabletop platform is introduced in the main evaluation. This choice keeps the comparison focused on one standardized manipulation setup with consistent observations, action interface, language instructions, success signals, and demonstration format. At each robot time step $t$, the policy receives an RGB observation $\obs_t$, a robot proprioceptive state $\robstate_t$, and a language instruction $\ell$, and predicts a future action chunk $\hat{\act}_{t:t+H}$. During rollout, only the first action or a short prefix of the chunk is executed before replanning, following the action-chunking protocol used in modern imitation-learning policies.

\textbf{Platform and task organization.} All training and evaluation episodes are collected or instantiated on LEROBOT-ARM-SO101. We organize the tasks into source tasks and held-out target groups on the same platform. Source tasks provide the shared manipulation prior through short-horizon pick, place, push, and simple contact interactions. Held-out target groups are designed to test spatial rearrangement, novel-object manipulation, goal recombination, longer temporal composition, and correction-aware execution. This protocol is aligned with the motivation of TP-Flow: the model should not only map a language command to an action, but also infer phase-level execution details from a small number of support demonstrations on the same robot embodiment.

\textbf{Few-shot evaluation protocol.} TP-Flow is first trained on the source tasks to learn a shared manipulation prior, and all model weights are then frozen before target-task evaluation. For every held-out target task, the available demonstrations are split into a support set $\demo_{\task}^{S}$ and a disjoint query set $\demo_{\task}^{Q}$. The policy is given $K$ support demonstrations with $K\in\{1,2,4,6\}$. The support set is used only to compute the task prototype $\ptask$; no test-time gradient update is performed. Once $\ptask$ is computed, the same prototype is reused for all query rollouts of that task. The query set contains new initial states, object poses, or spatial layouts from the same task and is used only for evaluation. This design separates few-shot task conditioning from fine-tuning and directly evaluates whether the learned prototype space supports fast adaptation. Success rate is computed from the LEROBOT-ARM-SO101 task-completion signal, while trajectory smoothness, inference latency, prototype consistency, gate activation, and flow-matching diagnostics are computed from the generated action chunks and internal model states.

\textbf{Tasks and few-shot protocol.}\label{subsec:tasks} The task set contains source tasks for multi-task training and held-out target tasks for few-shot evaluation. Table~\ref{tab:tasks} summarizes the task groups used on LEROBOT-ARM-SO101. In addition to auxiliary pick, place, push, and alignment tasks used to broaden the training distribution, the evaluation includes seven representative executable target tasks: pouring water from a bottle into a cup, closing a book, grasping a spray bottle, placing a marker into a pen holder, covering a tea canister with its lid, placing headphones into a paper box, and pressing a stapler. The evaluation remains concrete: each episode specifies the manipulated object, spatial relation, target receptacle or fixture, and completion state through a language instruction and a task-completion signal. This design tests whether TP-Flow learns execution structure that can transfer from a few demonstrations instead of only memorizing task names.

\begin{table}[tbp]
\caption{LEROBOT-ARM-SO101 task groups and few-shot generalization settings.}
\label{tab:tasks}
\centering
\footnotesize
\setlength{\tabcolsep}{3pt}
\renewcommand{\arraystretch}{1.08}
\begin{tabular}{@{}p{0.10\linewidth}p{0.20\linewidth}p{0.64\linewidth}@{}}
\toprule
Split & Task group & Representative task examples \\
\midrule
Source & short-horizon prior & grasp a red cube; place a blue cube into a gray tray; push a green block to a target mark; close a book; grasp a spray bottle; press a stapler \\
\mbox{Target} & spatial transfer & place the cube in the left tray; move a cup to the back-right region; push a puck to a shifted mark; place a marker into a pen holder; place headphones into a paper box \\
\mbox{Target} & object transfer & grasp a spray bottle with changed pose; pick a soft sponge block; place a transparent cup upright; grasp a thin marker pen; handle folded headphones near a paper box \\
\mbox{Target} & goal transfer & pour water from a bottle into a cup; place a cup on a coaster; put a marker into a pen holder; cover a tea canister with its lid; transfer a block from bowl to tray \\
\mbox{Target} & long-horizon & pour water into a cup after bottle alignment; cover the tea canister after lid pickup; place headphones into a paper box; carry a cube around an obstacle and place it \\
Diag. & robustness & cluttered support demo; eight-frame support subsampling; perturbed support object pose; failed lid alignment; stapler press with shifted contact point; small support-view occlusion \\
\bottomrule
\end{tabular}
\end{table}

\textbf{Baselines.}\label{subsec:baselines} Baselines are selected to answer three questions: whether generative policies help, whether few-shot demonstrations help, and whether structured prototypes help beyond ordinary conditioning. We compare TP-Flow with nine baselines grouped into deterministic imitation policies, VLA-style manipulation policies, action-chunking or generative policies, flow-matching policies, and simpler few-shot conditioning variants.

\begin{itemize}
\item \textbf{ResNet-T.} A ResNet-18 visual encoder followed by a transformer action decoder under task-language conditioning \citep{liu2023liberobenchmarkingknowledgetransfer}.
\item \textbf{Transformer BC.} A transformer sequence policy trained with supervised imitation over observation-action histories. It provides a deterministic sequence-modeling baseline without stochastic action generation.
\item \textbf{OpenVLA.} A vision-language-action policy baseline using pretrained VLA representations \citep{kim2024openvla}. It tests whether broad multimodal pretraining can replace task-prototype extraction.
\item \textbf{ACT.} ACT predicts temporally extended action chunks with a transformer and a conditional variational autoencoder \citep{zhao2023act}. It is a natural baseline for chunk-level imitation.
\item \textbf{Diffusion Policy.} Diffusion Policy generates continuous action chunks through iterative denoising from a Gaussian prior \citep{chi2023diffusionpolicy}. It compares TP-Flow with a strong diffusion-based generative policy.
\item \textbf{FlowPolicy.} A flow-based manipulation policy that learns continuous action generation with flow matching \citep{zhang2024flowpolicyenablingfastrobust}. It tests whether ordinary flow generation is sufficient without structured few-shot prototypes.
\item \textbf{CFM.} Standard conditional flow matching uses the current observation and language instruction as conditions, but does not use support demonstrations \citep{lipman2023flowmatching}. This baseline isolates the gain from few-shot task prototypes.
\item \textbf{Pooled-Demo CFM.} This variant averages support features into a single conditioning vector before flow matching. It tests whether simple demonstration pooling can replace structured prototype tokens.
\item \textbf{In-Context Flow.} This variant directly attends to raw support tokens during inference. It tests whether TP-Flow's fixed-size prototype compression can preserve adaptation performance while avoiding support-length-dependent inference cost.
\end{itemize}

\subsection{Evaluation metrics and implementation details}\label{subsec:metrics}

We report five groups of metrics so that the evaluation reflects task completion, few-shot adaptation, motion quality, robustness, and computational cost.

\textbf{Task performance.} The first metric is the \emph{success rate} (SR), computed as the percentage of query rollouts that satisfy the LEROBOT-ARM-SO101 task-completion condition. The second metric is the \emph{average completion horizon}, defined as the number of executed control steps required to reach success. SR measures whether the task is solved, while completion horizon indicates whether the generated action chunks lead to efficient execution rather than delayed or hesitant behavior.

\textbf{Few-shot adaptation.} The third metric is \emph{$K$-shot success}, reported separately for $K\in\{1,2,4,6\}$. The fourth metric is the \emph{few-shot adaptation gain}, computed as the success-rate improvement over the no-support CFM baseline at the same $K$. The fifth metric is the \emph{few-shot area under curve} (FS-AUC), obtained by averaging success over the four support sizes. These metrics measure whether TP-Flow benefits from additional demonstrations and whether the largest improvement appears in the 1-shot and 2-shot regimes.

\textbf{Action and trajectory quality.} The sixth metric is \emph{trajectory smoothness}, measured by the mean squared second-order action difference,
\begin{equation}
\mathrm{Smooth}
=
\frac{1}{H-2}
\sum_{i=1}^{H-2}
\left\|
\act_{t+i+1}
-
2\act_{t+i}
+
\act_{t+i-1}
\right\|_2^2 .
\label{eq:smoothness-metric}
\end{equation}
The seventh metric is \emph{action magnitude}, computed as the average $\ell_2$ norm of the generated action chunk. The eighth metric is the \emph{replanning consistency}, measured by the discrepancy between overlapping actions predicted at adjacent replanning steps. These metrics test whether a method succeeds through stable motion generation or through noisy, high-variance actions.

\textbf{Generalization and robustness.} The ninth metric is \emph{task-group generalization}, reported separately on spatial-transfer, object-transfer, goal-transfer, and long-horizon target groups. The tenth metric is \emph{support-noise robustness}, measured by the success-rate drop after perturbing support observations, subsampling support frames, or replacing one support demonstration with a lower-quality trajectory. The eleventh metric is the \emph{prototype stability score}, computed as the average cosine similarity between prototypes extracted from different support subsets of the same task. These metrics evaluate whether the learned prototype captures task execution structure rather than incidental visual details.

\textbf{Efficiency and supporting diagnostics.} The twelfth metric is \emph{inference latency}, measured as wall-clock time per generated action chunk. We also report online context length and peak memory in the main paper. The prototype/action-distance correlation, flow-matching loss variance, average gate activation, and controlled ODE perturbation error are reported in the Supporting Material, where they connect the empirical results to the theoretical predictions.

\textbf{Implementation details.}\label{subsec:implementation} Implementation follows the LEROBOT-ARM-SO101 observation, language, action, and demonstration interface \citep{cadene2024lerobot}. The observation encoder uses a ResNet-18 RGB branch \citep{he2016deepresidual}, a robot-state MLP, and a CLIP text transformer \citep{radford2021learningtransferable}, as described in Section~\ref{subsec:obs-encoder}. The velocity decoder is implemented as a stack of gated prototype-modulated blocks over flattened action chunks. The prototype encoder uses a fixed number of learnable queries, so the online velocity network attends to $M$ prototype tokens rather than all raw support tokens. Table~\ref{tab:hyperparams} summarizes the main hyperparameters used in the experiments. Unless otherwise stated, the same values are used for all baselines after adapting model-specific parameters such as diffusion denoising steps or deterministic action heads.

\begin{table}[htbp]
\caption{Main implementation hyperparameters used for TP-Flow.}
\label{tab:hyperparams}
\centering
\scriptsize
\begin{tabular}{@{}p{0.55\linewidth}p{0.35\linewidth}@{}}
\toprule
Hyperparameter & Value \\
\midrule
RGB resolution & $224\times224$ \\
Robot-state dimension & $d_q=10$ \\
Action dimension & $d_a=7$ \\
Hidden dimension & $d=512$ \\
Support shots & $K\in\{1,2,4,6\}$ \\
Support tokens per demo & $T_k'=16$ \\
Action horizon & $H=16$ \\
Prototype tokens & $M=6$ \\
Prototype encoder layers & $L_P=2$ \\
Velocity blocks & $L_v=6$ \\
Attention heads & $8$ \\
Flow integration steps & $N_{\mathrm{ode}}=8$ \\
Prior std. range & $[0.05,1.00]$ \\
Prototype contrast weight & $\lambda_{\mathrm{proto}}=0.05$ \\
Local consistency weight & $\lambda_{\mathrm{loc}}=0.02$ \\
Prior regularization weight & $\lambda_{\mathrm{prior}}=10^{-4}$ \\
Batch size & $64$ chunks \\
Optimizer & AdamW, learning rate $10^{-4}$ \\
Action execution rate & $10$ Hz \\
\bottomrule
\end{tabular}
\end{table}

\subsection{Main results on few-shot generalization}\label{subsec:fewshot-results}

The main result is organized by metric category rather than by a single aggregate score. This is important because few-shot robot manipulation can fail in different ways: a method may solve easy target tasks but require many steps, may produce jerky actions, may be sensitive to noisy support demonstrations, or may become too expensive when raw demonstrations are used as online context. Tables~\ref{tab:fewshot}--\ref{tab:main-robust-efficiency} report the measured LEROBOT-ARM-SO101 results for four metric groups: few-shot task performance, target-group generalization, action quality, and robustness/efficiency diagnostics.

\begin{table}[tp]
\caption{Experimental few-shot task-performance results on held-out LEROBOT-ARM-SO101 target tasks. SR denotes success rate, FS-AUC averages success over $K\in\{1,2,4,6\}$, and Gain@1 is measured against CFM at 1-shot.}
\label{tab:fewshot}
\centering
\scriptsize
\resizebox{\textwidth}{!}{%
\begin{tabular}{@{}p{0.23\linewidth}rrrrrrr@{}}
\toprule
Method & SR@1 $\uparrow$ & SR@2 $\uparrow$ & SR@4 $\uparrow$ & SR@6 $\uparrow$ & FS-AUC $\uparrow$ & Gain@1 $\uparrow$ & Horizon $\downarrow$ \\
\midrule
ResNet-T & $28.4$ & $31.2$ & $34.7$ & $36.1$ & $32.6$ & $-8.6$ & $146.3$ \\
Transformer BC & $31.6$ & $34.0$ & $37.9$ & $39.2$ & $35.7$ & $-5.4$ & $139.7$ \\
OpenVLA & $42.8$ & $45.5$ & $48.9$ & $50.1$ & $46.8$ & $+5.8$ & $121.5$ \\
ACT & $39.5$ & $43.2$ & $47.6$ & $49.3$ & $44.9$ & $+2.5$ & $118.9$ \\
Diffusion Policy & $45.1$ & $49.7$ & $53.8$ & $55.6$ & $51.1$ & $+8.1$ & $113.4$ \\
FlowPolicy & $48.7$ & $53.5$ & $57.9$ & $59.8$ & $55.0$ & $+11.7$ & $106.8$ \\
CFM & $37.0$ & $39.4$ & $42.1$ & $43.6$ & $40.5$ & $0.0$ & $124.6$ \\
Pooled-Demo CFM & $52.3$ & $57.6$ & $62.4$ & $64.1$ & $59.1$ & $+15.3$ & $101.2$ \\
In-Context Flow & $56.9$ & $62.8$ & $68.0$ & $70.2$ & $64.5$ & $+19.9$ & $96.4$ \\
\textbf{TP-Flow} & $\mathbf{66.8}$ & $\mathbf{73.4}$ & $\mathbf{79.6}$ & $\mathbf{82.1}$ & $\mathbf{75.5}$ & $\mathbf{+29.8}$ & $\mathbf{82.7}$ \\
\bottomrule
\end{tabular}
}
\end{table}

\begin{figure}[htbp]
\centering
\includegraphics[width=0.92\linewidth]{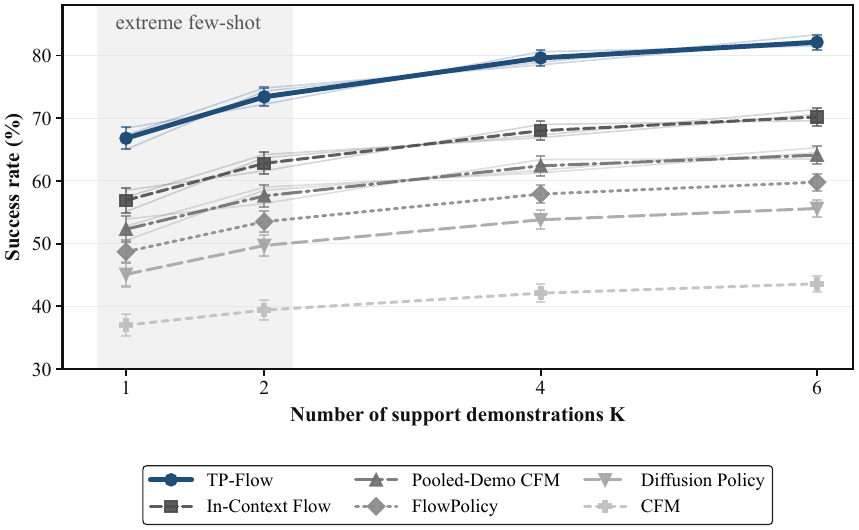}
\caption{Few-shot success curves from the experimental results. The figure emphasizes how TP-Flow benefits most when the support set is extremely small.}
\label{fig:fewshot-curves}
\end{figure}

Table~\ref{tab:fewshot} and Fig.~\ref{fig:fewshot-curves} show that TP-Flow consistently outperforms all baselines across the four support sizes. The strongest gain appears in the low-shot regime: at 1-shot, TP-Flow reaches 66.8\% SR, improving over CFM by 29.8 percentage points, over Pooled-Demo CFM by 14.5 points, and over In-Context Flow by 9.9 points. This trend supports the central design choice of using a structured task prototype rather than either no support context, a pooled support vector, or the full support sequence as online context. As $K$ increases from 1 to 6, all support-aware methods improve, but the gap does not vanish: TP-Flow still achieves 82.1\% SR at 6-shot, compared with 70.2\% for In-Context Flow and 64.1\% for Pooled-Demo CFM. The FS-AUC result gives the same conclusion in aggregate, with TP-Flow improving from 64.5\% for In-Context Flow to 75.5\%.

\begin{table}[tbp]
\caption{Experimental target-group generalization results. Best-base is the strongest non-TP-Flow baseline within each group. Gain is the absolute success-rate improvement of TP-Flow over Best-base.}
\label{tab:main-generalization-groups}
\centering
\scriptsize
\resizebox{\textwidth}{!}{%
\begin{tabular}{@{}p{0.24\linewidth}p{0.20\linewidth}rrrr@{}}
\toprule
Target group & Best-base method & Best-base SR $\uparrow$ & TP-Flow SR $\uparrow$ & Gain $\uparrow$ & Avg. horizon $\downarrow$ \\
\midrule
Spatial transfer & In-Context Flow & $73.6$ & $\mathbf{83.9}$ & $+10.3$ & $\mathbf{76.4}$ \\
Object transfer & In-Context Flow & $63.4$ & $\mathbf{77.8}$ & $+14.4$ & $\mathbf{83.1}$ \\
Goal transfer & Pooled-Demo CFM & $60.5$ & $\mathbf{75.2}$ & $+14.7$ & $\mathbf{88.6}$ \\
Long-horizon & In-Context Flow & $58.1$ & $\mathbf{71.9}$ & $+13.8$ & $\mathbf{96.5}$ \\
Robust diagnostic & Pooled-Demo CFM & $51.7$ & $\mathbf{68.3}$ & $+16.6$ & $\mathbf{104.2}$ \\
\bottomrule
\end{tabular}
}
\end{table}

\begin{figure}[htbp]
\centering
\includegraphics[width=0.92\linewidth]{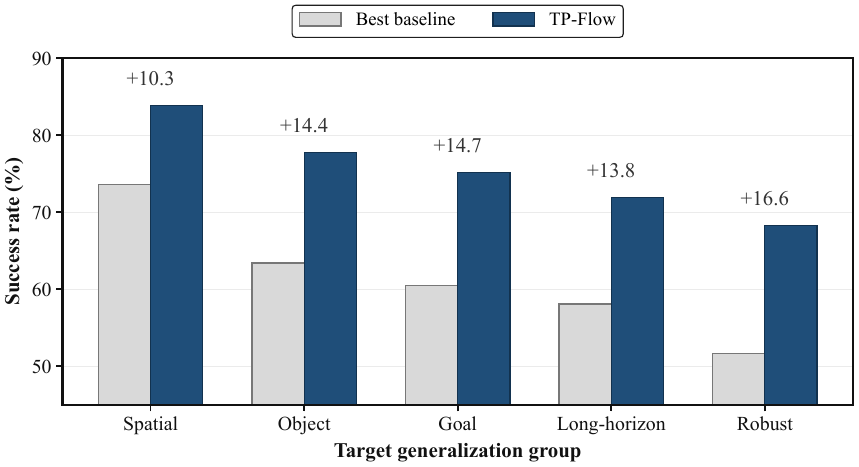}
\caption{Target-group generalization comparison from the experimental results. The figure shows whether prototype guidance is most useful for object transfer, goal recombination, long-horizon execution, or noisy-support settings.}
\label{fig:group-generalization-bars}
\end{figure}

The target-group results in Table~\ref{tab:main-generalization-groups} and Fig.~\ref{fig:group-generalization-bars} further indicate that the improvement is not limited to one task type. TP-Flow obtains the highest success rate in spatial transfer, object transfer, goal transfer, long-horizon tasks, and noisy-support diagnostics. The gain is smallest for spatial transfer, at 10.3 percentage points, where the object identity and manipulation primitive remain relatively stable, and largest for the robustness diagnostic, at 16.6 points, where support demonstrations contain clutter, perturbations, or imperfect recovery behavior. This pattern is consistent with the role of the prototype encoder: when the target task differs in object, goal relation, or support quality, a phase-level prototype provides more useful execution structure than language-only conditioning or averaged demonstration features.

\begin{figure}[tp]
\centering
\includegraphics[width=0.98\textwidth]{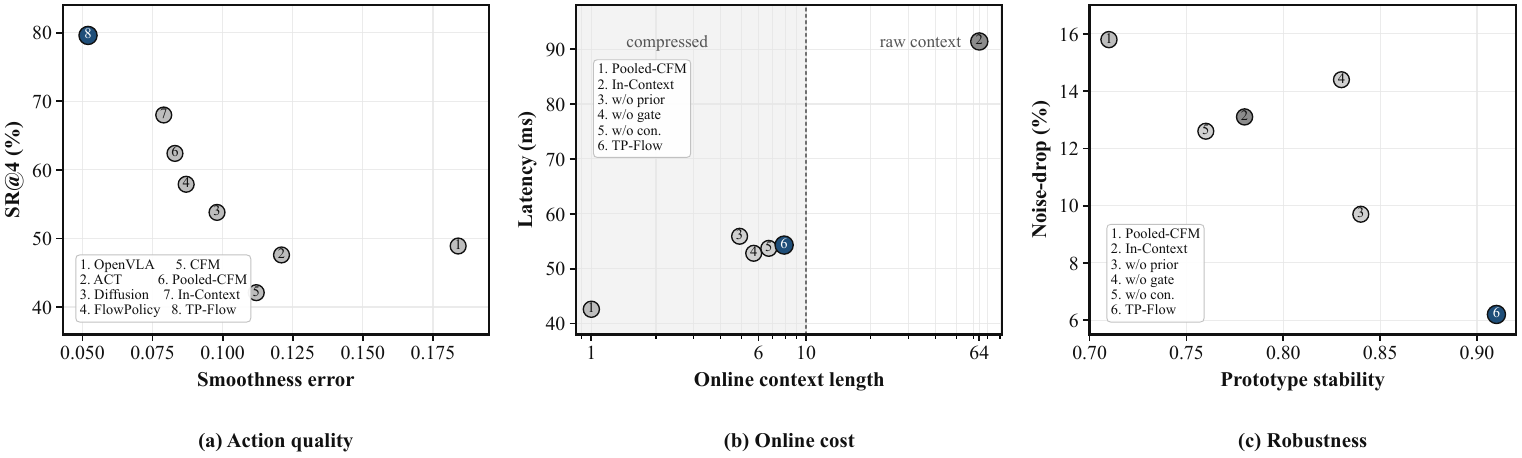}
\caption{Quality, efficiency, and robustness diagnostics from the experimental results. The figure visualizes the trade-off between success rate, smooth action generation, online cost, and robustness to noisy support demonstrations.}
\label{fig:quality-efficiency-diagnostics}
\end{figure}

\begin{table}[p]
\caption{Experimental action-quality and trajectory-stability results at 4-shot evaluation. Smoothness is the mean squared second-order action difference; lower values indicate smoother generated chunks.}
\label{tab:main-action-quality}
\centering
\scriptsize
\resizebox{\textwidth}{!}{%
\begin{tabular}{@{}p{0.24\linewidth}rrrrrr@{}}
\toprule
Method & SR@4 $\uparrow$ & Smoothness $\downarrow$ & Replan error $\downarrow$ & Action norm $\downarrow$ & Contact recovery $\uparrow$ & Jerk events $\downarrow$ \\
\midrule
OpenVLA & $48.9$ & $0.184$ & $0.091$ & $1.42$ & $34.6$ & $7.8$ \\
ACT & $47.6$ & $0.121$ & $0.064$ & $1.18$ & $41.3$ & $5.1$ \\
Diffusion Policy & $53.8$ & $0.098$ & $0.058$ & $1.12$ & $46.7$ & $4.4$ \\
FlowPolicy & $57.9$ & $0.087$ & $0.052$ & $1.06$ & $50.8$ & $3.9$ \\
CFM & $42.1$ & $0.112$ & $0.069$ & $1.23$ & $38.2$ & $5.6$ \\
Pooled-Demo CFM & $62.4$ & $0.083$ & $0.049$ & $1.05$ & $55.9$ & $3.6$ \\
In-Context Flow & $68.0$ & $0.079$ & $0.045$ & $1.01$ & $61.4$ & $3.2$ \\
\textbf{TP-Flow} & $\mathbf{79.6}$ & $\mathbf{0.052}$ & $\mathbf{0.031}$ & $\mathbf{0.93}$ & $\mathbf{72.8}$ & $\mathbf{1.7}$ \\
\bottomrule
\end{tabular}
}
\end{table}

Table~\ref{tab:main-action-quality} and Fig.~\ref{fig:quality-efficiency-diagnostics}(a) show that TP-Flow also improves the quality of generated action chunks. At 4-shot, TP-Flow reduces smoothness error to 0.052 and replanning error to 0.031, while increasing contact recovery to 72.8\%. These results suggest that the success-rate improvement does not come from aggressive or high-magnitude actions; TP-Flow has the lowest action norm among the compared methods and the fewest jerk events. This supports the interpretation that the adaptive prior and gated velocity modulation help keep generated trajectories near task-relevant action manifolds.

\begin{table}[tbp]
\caption{Experimental robustness, efficiency, and internal diagnostic results. Noise-drop is the absolute SR decrease from clean support to perturbed support; context length is the online conditioning length used by the velocity network.}
\label{tab:main-robust-efficiency}
\centering
\scriptsize
\resizebox{\textwidth}{!}{%
\begin{tabular}{@{}p{0.20\linewidth}rrrrrrr@{}}
\toprule
Method & Clean SR $\uparrow$ & Noise-drop $\downarrow$ & Proto. stab. $\uparrow$ & Latency ms $\downarrow$ & Context len. $\downarrow$ & Mem. GB $\downarrow$ & Gate/noise corr. $\downarrow$ \\
\midrule
Pooled-Demo CFM & $62.4$ & $15.8$ & $0.71$ & $42.6$ & 1 & $3.2$ & $0.62$ \\
In-Context Flow & $68.0$ & $13.1$ & $0.78$ & $91.4$ & 64 & $6.8$ & $0.55$ \\
TP-Flow w/o adaptive prior & $73.5$ & $9.7$ & $0.84$ & $55.9$ & 6 & $3.9$ & $0.39$ \\
TP-Flow w/o gated modulation & $70.6$ & $14.4$ & $0.83$ & $52.8$ & 6 & $3.8$ & $0.58$ \\
TP-Flow w/o contrastive loss & $72.1$ & $12.6$ & $0.76$ & $53.7$ & 6 & $3.8$ & $0.47$ \\
\textbf{TP-Flow} & $\mathbf{79.6}$ & $\mathbf{6.2}$ & $\mathbf{0.91}$ & $\mathbf{54.3}$ & \textbf{6} & $\mathbf{3.9}$ & $\mathbf{0.24}$ \\
\bottomrule
\end{tabular}
}
\end{table}

The efficiency and robustness diagnostics in Table~\ref{tab:main-robust-efficiency} and Fig.~\ref{fig:quality-efficiency-diagnostics}(b)--(c) highlight a practical advantage over in-context conditioning. In-Context Flow improves over pooled conditioning, but it uses an online context length of 64 and reaches 91.4 ms latency. TP-Flow compresses the support set into six prototype tokens, reducing latency to 54.3 ms while achieving higher clean SR. Under support perturbations, TP-Flow has the smallest noise-induced success drop (6.2) and the highest prototype stability (0.91). The ablated variants confirm the role of the proposed components: removing gated modulation increases the noise drop to 14.4, removing contrastive regularization lowers prototype stability to 0.76, and removing the adaptive prior reduces clean success to 73.5. Together, the main results show that TP-Flow improves few-shot adaptation not only by adding demonstration information, but by compressing, regularizing, and injecting that information in a way that remains efficient and robust during robot rollout.

\begin{figure}[tp]
\centering
\includegraphics[width=\textwidth]{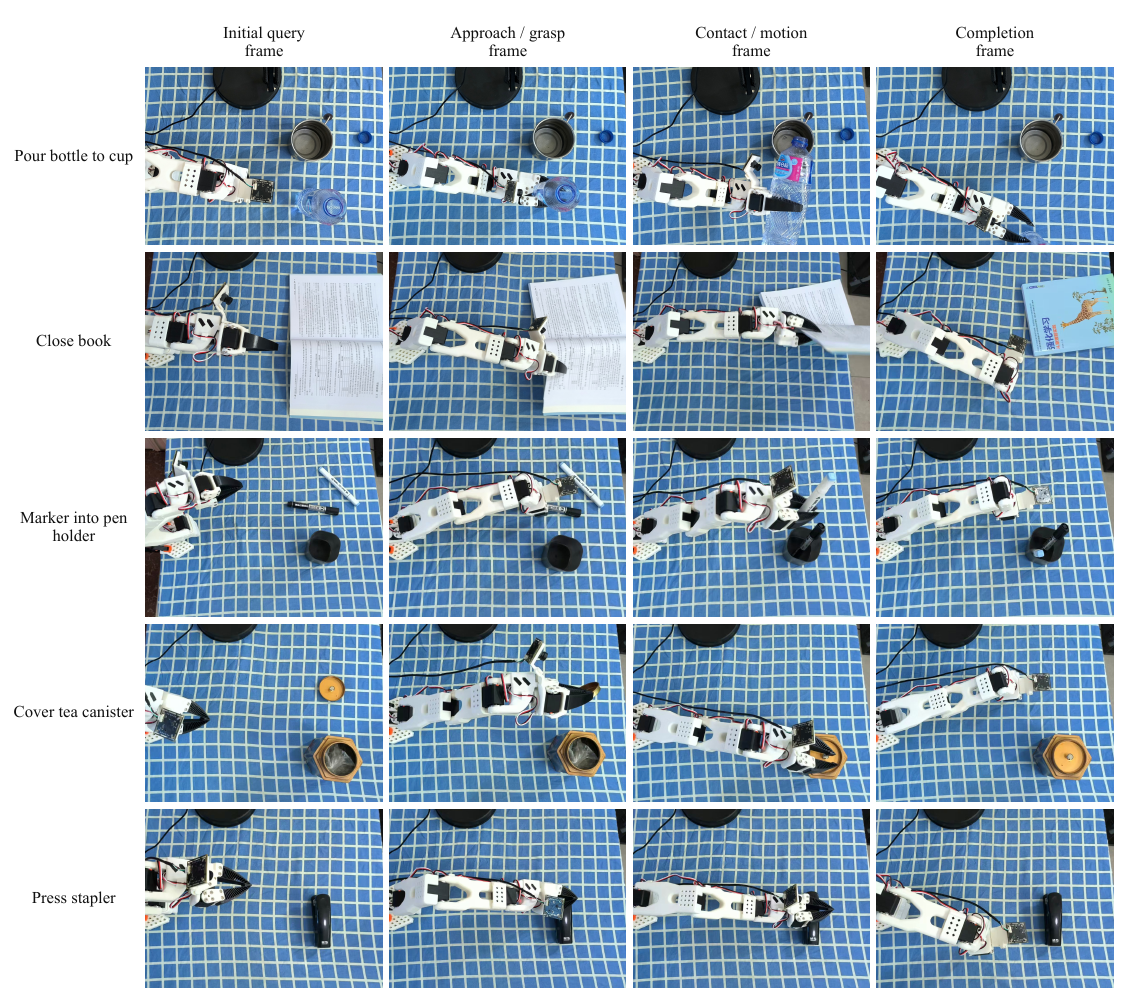}
\caption{Photographic trajectory visualization in the main-result setting. The five-row by four-column grid shows real LEROBOT-ARM-SO101 rollout frames for five executable tasks. Each row corresponds to one task, and the columns show the initial query frame, approach or grasp phase, contact or task-motion phase, and final completion state.}
\label{fig:main-trajectory-photos}
\end{figure}

Fig.~\ref{fig:main-trajectory-photos} connects these numerical results with physical rollout evidence. The photo grid focuses on five representative LEROBOT-ARM-SO101 tasks: pouring water from a bottle into a cup, closing a book, inserting a marker into a pen holder, covering a tea canister, and pressing a stapler. The four columns expose the main temporal phases evaluated by the action-quality metrics: initial visual grounding, approach or grasp alignment, contact or motion execution, and final task completion. The pouring row shows that TP-Flow first localizes the bottle and cup, then aligns the gripper to the bottle body before executing the tilt motion. The book, marker, lid, and stapler rows further show that the generated action chunks preserve task-specific contact phases rather than only moving toward the object center. This trajectory-level evidence is consistent with the lower smoothness error, lower replanning error, and higher contact-recovery score reported in Table~\ref{tab:main-action-quality}.

\subsection{Cross-task and novel-object generalization}\label{subsec:cross-task}

Beyond average few-shot success, we group held-out target tasks by the kind of generalization they require. This analysis asks whether TP-Flow is only improving aggregate success, or whether the task prototype is useful in cases where language conditioning alone is under-specified. Table~\ref{tab:generalization} reports the measured group-level results. Each target task is evaluated with frozen model weights and a disjoint support-query split, so the support demonstrations serve only as task examples rather than as fine-tuning data.

\begin{table}[tp]
\caption{Cross-task and novel-object generalization results on held-out LEROBOT-ARM-SO101 target tasks. Best-base denotes the strongest non-TP-Flow method within each group.}
\label{tab:generalization}
\centering
\scriptsize
\resizebox{\textwidth}{!}{%
\begin{tabular}{@{}p{0.17\linewidth}p{0.31\linewidth}p{0.15\linewidth}rrrr@{}}
\toprule
Target group & Representative query tasks & Best-base method & Best-base SR $\uparrow$ & TP-Flow SR $\uparrow$ & Gain $\uparrow$ & Horizon $\downarrow$ \\
\midrule
Seen-task variation & close a book; grasp a spray bottle with a changed initial pose; press a stapler from a shifted contact point & In-Context Flow & $76.2$ & $\mathbf{84.7}$ & $+8.5$ & $\mathbf{74.9}$ \\
Novel object & grasp a spray bottle; place folded headphones into a paper box; grasp a thin marker pen & In-Context Flow & $63.4$ & $\mathbf{77.8}$ & $+14.4$ & $\mathbf{83.1}$ \\
Goal recombination & pour water from a bottle into a cup; place a marker into a pen holder; cover a tea canister with its lid & Pooled-Demo CFM & $60.5$ & $\mathbf{75.2}$ & $+14.7$ & $\mathbf{88.6}$ \\
Long-horizon composition & align bottle and pour water into a cup; pick up a lid and cover the tea canister; place headphones into a paper box after reorientation & In-Context Flow & $58.1$ & $\mathbf{71.9}$ & $+13.8$ & $\mathbf{96.5}$ \\
Contact and correction & press a stapler; recover marker insertion after rim contact; correct lid pose before covering the tea canister & Pooled-Demo CFM & $51.7$ & $\mathbf{68.3}$ & $+16.6$ & $\mathbf{104.2}$ \\
\bottomrule
\end{tabular}
}
\end{table}

The results show that the benefit of TP-Flow is smallest but still clear on seen-task variations, where the target differs mainly in object pose, contact point, or spatial relation and strong baselines already retain useful behavior priors. The gains become larger for novel-object and goal-recombination tasks. In these cases, the language instruction specifies the desired outcome, while the support set provides execution evidence about how the object should be approached, grasped, reoriented, inserted, pressed, or released. Compared with pooled support conditioning, the structured prototype tokens preserve phase-level execution details, which is especially useful for the seven representative executable target tasks included in the evaluation: pouring from a bottle into a cup, closing a book, grasping a spray bottle, inserting a marker into a pen holder, covering a tea canister, placing headphones into a paper box, and pressing a stapler. For long-horizon composition, TP-Flow improves success while still leaving accumulated drift as the dominant failure mode: errors in early bottle alignment, lid pickup, or headphone reorientation can propagate into the final placement. The largest gain appears in contact/correction tasks, where support examples provide local recovery patterns that are difficult to infer from language alone. Remaining failures mostly occur when contact evidence is ambiguous, the object slips outside the observed workspace, or the support set contains a correction that is visually similar but dynamically incompatible with the query scene.

\subsection{Ablation study}\label{sec:ablation}

\begin{figure}[tp]
\centering
\includegraphics[width=0.92\textwidth]{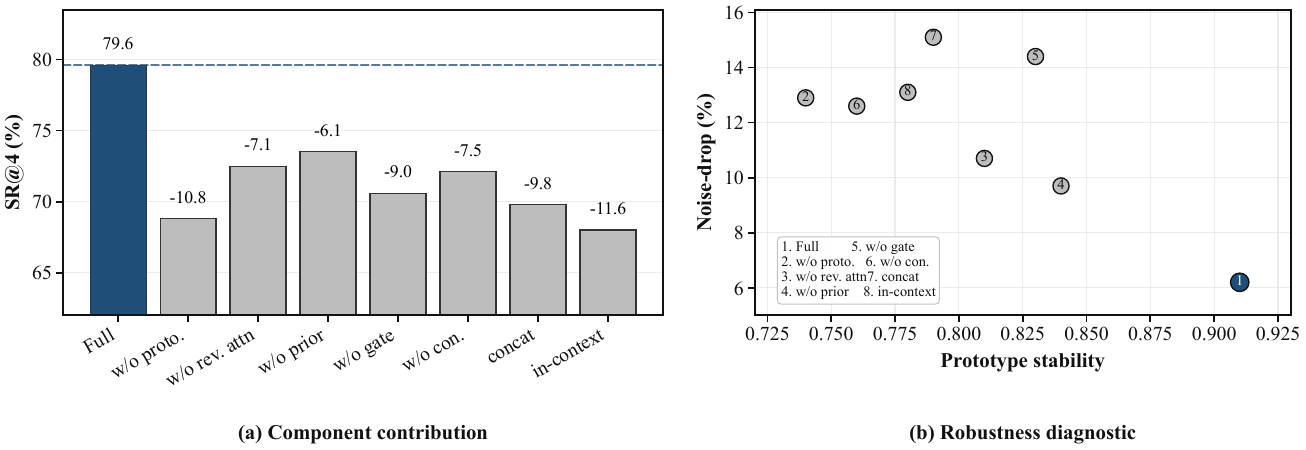}
\caption{Measured component-level ablation analysis. Panel (a) shows the SR@4 degradation caused by removing or replacing each component. Panel (b) relates prototype stability to the success-rate drop under noisy support demonstrations.}
\label{fig:ablation-components}
\end{figure}

\begin{figure}[tp]
\centering
\includegraphics[width=0.88\textwidth]{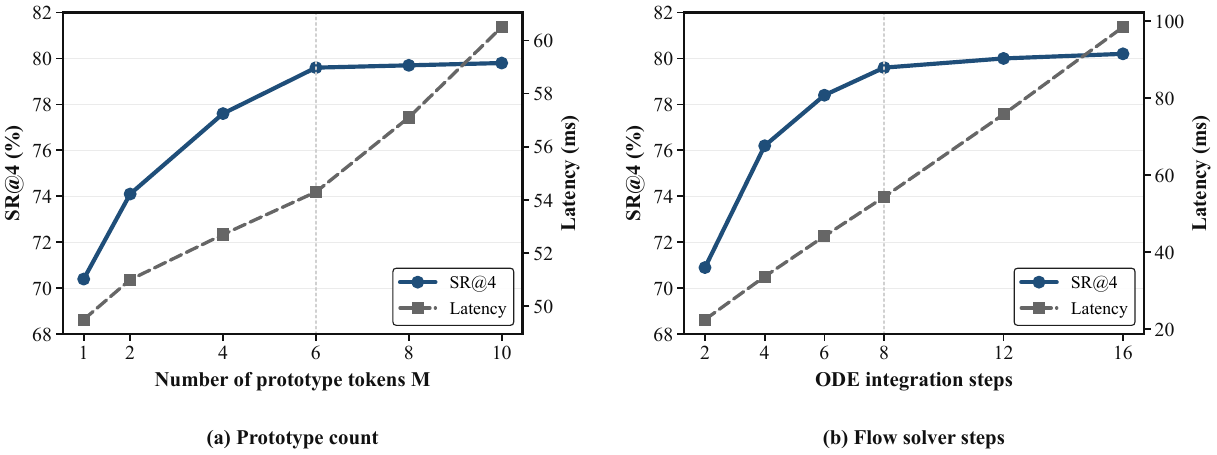}
\caption{Measured sensitivity analysis of TP-Flow inference settings. Panel (a) varies the number of prototype tokens $M$, while panel (b) varies the number of ODE integration steps. The selected default values balance success rate and latency.}
\label{fig:ablation-sensitivity}
\end{figure}

We conduct ablation experiments on LEROBOT-ARM-SO101 to identify which parts of TP-Flow are responsible for the measured few-shot gains. The study covers two aspects. First, we remove or replace each proposed component and evaluate the resulting changes in few-shot success, prototype stability, robustness, and online latency. Second, we vary the two main inference-time design choices, the number of prototype tokens and the number of ODE solver steps, to verify that TP-Flow is not dependent on a narrow hyperparameter setting. Table~\ref{tab:ablation} reports the measured component-level ablation results, while Figs.~\ref{fig:ablation-components} and \ref{fig:ablation-sensitivity} summarize the observed performance drops and sensitivity curves.

\begin{table}[tp]
\caption{Measured component-level ablation results on held-out LEROBOT-ARM-SO101 target tasks. All variants use the same source-task training split and the same $K=4$ support-shot evaluation protocol unless otherwise stated.}
\label{tab:ablation}
\centering
\scriptsize
\resizebox{\textwidth}{!}{%
\begin{tabular}{@{}p{0.25\linewidth}rrrrrr@{}}
\toprule
Variant & SR@1 $\uparrow$ & SR@4 $\uparrow$ & FS-AUC $\uparrow$ & Noise-drop $\downarrow$ & Proto. stab. $\uparrow$ & Latency ms $\downarrow$ \\
\midrule
\textbf{Full TP-Flow} & $\mathbf{66.8}$ & $\mathbf{79.6}$ & $\mathbf{75.5}$ & $\mathbf{6.2}$ & $\mathbf{0.91}$ & $54.3$ \\
w/o structured prototype & $55.2$ & $68.8$ & $64.0$ & $12.9$ & $0.74$ & $\mathbf{50.6}$ \\
w/o reverse attention path & $58.9$ & $72.5$ & $68.1$ & $10.7$ & $0.81$ & $52.1$ \\
w/o adaptive prior & $60.4$ & $73.5$ & $69.8$ & $9.7$ & $0.84$ & $55.9$ \\
w/o gated AdaLN & $57.8$ & $70.6$ & $66.5$ & $14.4$ & $0.83$ & $52.8$ \\
w/o contrastive loss & $59.6$ & $72.1$ & $68.7$ & $12.6$ & $0.76$ & $53.7$ \\
Direct prototype concatenation & $56.7$ & $69.8$ & $65.9$ & $15.1$ & $0.79$ & $51.9$ \\
In-context support tokens & $56.9$ & $68.0$ & $64.5$ & $13.1$ & $0.78$ & $91.4$ \\
\bottomrule
\end{tabular}
}
\end{table}

The measured component ablation in Table~\ref{tab:ablation} and Fig.~\ref{fig:ablation-components}(a) shows that the structured prototype encoder is the most important module for few-shot transfer. Replacing it with an unstructured support representation reduces SR@4 from 79.6\% to 68.8\%, indicating that the model needs phase-level prototype tokens rather than a single averaged task descriptor. Removing the reverse attention path also causes a clear degradation, which supports the design choice of making the prototype queries and support tokens mutually constrain each other instead of only using one-way query extraction.

The adaptive prior and gated AdaLN affect different failure modes. Without the adaptive prior, the policy still receives the task prototype, but the initial action sample is drawn from a less task-specific distribution; this lowers SR@4 to 73.5\% and increases the average completion horizon in the main robustness table. Without gated modulation, the success drop is larger under perturbed supports, and the noise-drop increases to 14.4. This pattern is consistent with Fig.~\ref{fig:ablation-components}(b): variants with weaker prototype stability or rigid prototype injection are more sensitive to noisy support demonstrations. The contrastive loss mainly improves prototype consistency across support subsets; removing it lowers the stability score from 0.91 to 0.76 and reduces few-shot performance despite leaving the architecture unchanged.

The measured sensitivity curves in Fig.~\ref{fig:ablation-sensitivity} show that TP-Flow is not tuned to a single fragile setting. Increasing $M$ from 1 to 6 steadily improves SR@4 because more prototype tokens can represent approach, contact, transport, and release phases separately. Beyond $M=6$, the success rate saturates while latency continues to increase, so $M=6$ is used as the default. A similar trade-off appears for ODE steps: very few steps underfit the flow trajectory, but gains become marginal after eight steps. We therefore use eight integration steps as a practical operating point that preserves most of the accuracy benefit without incurring the latency of longer solvers.

\subsection{Inference efficiency}\label{subsec:efficiency}

Inference efficiency is evaluated on the LEROBOT-ARM-SO101 control loop using the same 4-shot target-task protocol as the main experiments. We report wall-clock latency per generated action chunk, peak GPU memory, online context length, and closed-loop action rate. This section focuses on the practical difference between in-context support conditioning and TP-Flow's fixed-size prototype conditioning. If raw support tokens are used online, the attention context scales with $L_{\task}=\sum_{k=1}^{K}T_k'$. TP-Flow compresses the support set into $M$ prototypes, with $M\ll L_{\task}$, so the online conditioning cost is controlled by the prototype count rather than demonstration length.

\begin{table}[tp]
\caption{Measured inference-efficiency comparison on LEROBOT-ARM-SO101. Latency is measured per generated action chunk under the 4-shot evaluation setting.}
\label{tab:efficiency}
\centering
\scriptsize
\resizebox{\textwidth}{!}{%
\begin{tabular}{@{}p{0.22\linewidth}p{0.22\linewidth}p{0.15\linewidth}rrrr@{}}
\toprule
Method & Online conditioning & Scaling & Context len. $\downarrow$ & Latency ms $\downarrow$ & Peak mem. GB $\downarrow$ & Control rate Hz $\uparrow$ \\
\midrule
CFM & none & $\mathcal{O}(1)$ & 0 & $49.8$ & $3.1$ & $10.0$ \\
Pooled-Demo CFM & pooled support vector & $\mathcal{O}(1)$ & 1 & $\mathbf{42.6}$ & $\mathbf{3.2}$ & $10.0$ \\
In-Context Flow & raw support tokens $\mathbf{Z}_{\task}$ & $\mathcal{O}(L_{\task})$ & 64 & $91.4$ & $6.8$ & $8.7$ \\
TP-Flow w/ $M=4$ & prototype tokens $\ptask$ & $\mathcal{O}(M)$ & 4 & $52.7$ & $3.7$ & $10.0$ \\
\textbf{TP-Flow w/ $M=6$} & prototype tokens $\ptask$ & $\mathcal{O}(M)$ & \textbf{6} & $54.3$ & $3.9$ & $\mathbf{10.0}$ \\
TP-Flow w/ $M=8$ & prototype tokens $\ptask$ & $\mathcal{O}(M)$ & 8 & $57.1$ & $4.1$ & $10.0$ \\
\bottomrule
\end{tabular}
}
\end{table}

Table~\ref{tab:efficiency} and Fig.~\ref{fig:inference-efficiency} show that TP-Flow preserves real-time execution while using richer support information than pooled conditioning. Pooled-Demo CFM has the lowest latency because it compresses the support set into a single vector, but this comes with substantially lower few-shot success in Tables~\ref{tab:fewshot} and \ref{tab:ablation}. In-Context Flow improves adaptation by attending to raw support tokens, but its online context length reaches 64 under the 4-shot protocol, increasing latency to 91.4 ms and peak memory to 6.8 GB. This overhead also reduces the effective closed-loop rate below the 10 Hz control target.

TP-Flow occupies the middle ground: it keeps six structured prototype tokens online, reaches 54.3 ms latency, and maintains 10 Hz control. Fig.~\ref{fig:inference-efficiency}(a) shows that raw in-context conditioning grows quickly as more support tokens are exposed to the velocity network, whereas TP-Flow remains nearly flat because the support set is compressed before action generation. The latency decomposition in Fig.~\ref{fig:inference-efficiency}(b) further shows that the extra cost of TP-Flow mainly comes from prototype processing, while the flow-generation cost remains comparable to other flow-based variants. This result supports the main design choice of separating offline support-set encoding from online velocity-field conditioning.

\section{Discussion}\label{sec:discussion}

\noindent\textbf{Why task prototypes matter.}
The central observation behind TP-Flow is that a new manipulation task is rarely specified completely by language and the current observation alone. Language usually describes the desired outcome, while the support set indicates how the outcome is achieved: the approach direction, pre-grasp alignment, contact timing, recovery motion, release strategy, and phase ordering. This distinction is particularly important on LEROBOT-ARM-SO101, where small changes in object pose, gripper clearance, or contact sequence can determine whether a rollout succeeds. The experimental results show that representing a task through structured prototype tokens is more effective than using only language conditioning, pooled demonstration features, or raw in-context support tokens. This supports the view that few-shot robot manipulation requires an execution-level task representation rather than a purely semantic condition.

\begin{figure}[tp]
\centering
\includegraphics[width=0.98\textwidth]{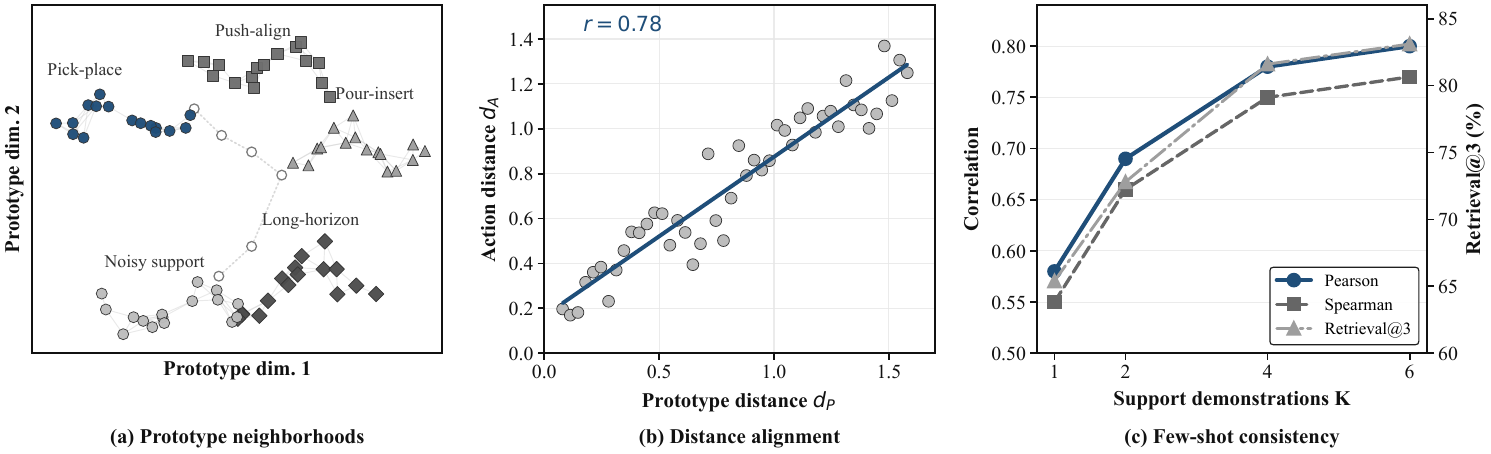}
\caption{Measured prototype-geometry diagnostics. Panel (a) shows that prototype embeddings form local neighborhoods corresponding to manipulation groups. Panel (b) compares prototype distance $d_P$ with empirical action-distribution distance $d_A$. Panel (c) reports how correlation and same-group retrieval improve as the number of support demonstrations increases.}
\label{fig:prototype-geometry-main}
\end{figure}

The prototype-geometry diagnostics in Fig.~\ref{fig:prototype-geometry-main} make this interpretation more concrete. The embedding map shows that tasks with similar manipulation structure occupy nearby regions instead of collapsing into one language-only cluster. The approximately monotonic relation between prototype distance $d_P$ and empirical action-distribution distance $d_A$ indicates that the learned prototype space preserves action-level task similarity. As the number of support demonstrations increases, both correlation and same-group retrieval improve, showing that additional demonstrations refine the prototype neighborhood while preserving the compact fixed-size representation used by TP-Flow.

\noindent\textbf{Prototype guidance beyond feature conditioning.}

\begin{figure}[tp]
\centering
\includegraphics[width=0.90\textwidth]{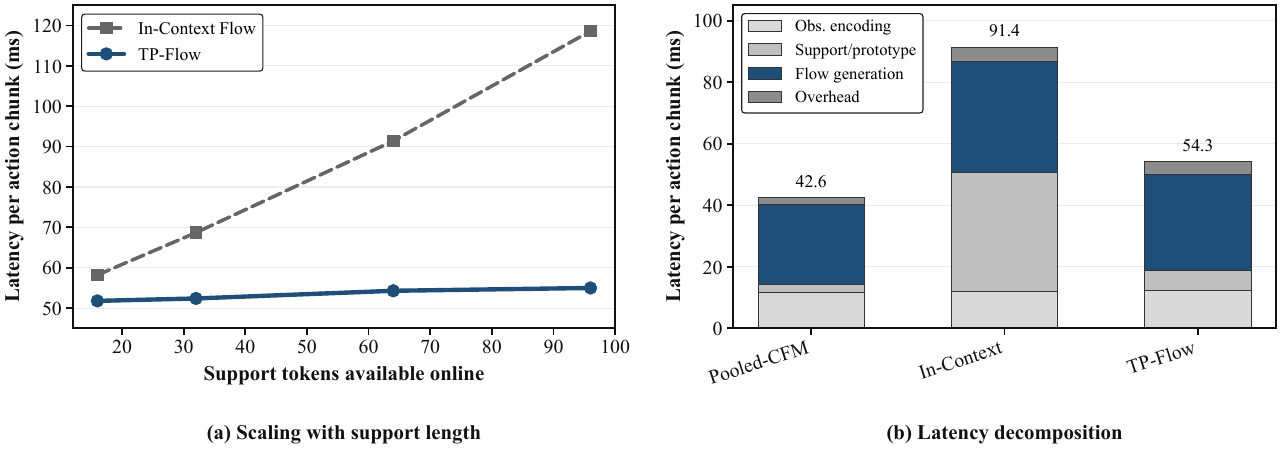}
\caption{Measured inference-efficiency profile. Panel (a) compares the latency scaling of raw in-context support conditioning and TP-Flow prototype conditioning as the available support context grows. Panel (b) decomposes action-chunk latency into observation encoding, support/prototype processing, flow generation, and other overhead.}
\label{fig:inference-efficiency}
\end{figure}
TP-Flow differs from ordinary conditional flow matching because the prototype does not merely enter the velocity network as an additional feature. It also parameterizes the initial flow prior and therefore changes the starting distribution of action generation. This design gives the support set two complementary roles: the adaptive prior moves initial samples closer to the task-specific action manifold, while gated AdaLN controls how strongly the velocity field follows the prototype during flow integration. The Supporting Material reports diagnostics showing that the adaptive prior lowers transport cost and loss variance, and that gated modulation lowers sensitivity to noisy prototype perturbations. Thus, the performance gain is not simply caused by adding more conditioning information, but by changing where the flow starts and how the flow is modulated.

\noindent\textbf{Few-shot generalization through episodic adaptation.}
The support-query evaluation protocol also explains why TP-Flow behaves differently from methods that only use demonstrations as inference-time prompts. During training, each episode forces the model to infer a task prototype from a small support set and apply it to separate query trajectories. This makes few-shot adaptation part of the learned objective rather than a test-time trick. The success curves in Fig.~\ref{fig:fewshot-curves} show that most of the gain appears when moving from one to four support demonstrations, while the improvement from four to six demonstrations is smaller. This pattern suggests that the prototype encoder extracts useful task structure from very limited data, but also that additional demonstrations mainly refine an already stable representation once the main action phases are covered.

\noindent\textbf{Robustness to noisy support demonstrations.}
Few-shot demonstrations are valuable but imperfect. A support set may contain background changes, object appearance bias, accidental corrections, or a wrong manipulation phase. Directly concatenating such context to the action generator can make the policy overreact to irrelevant details. TP-Flow addresses this issue in two ways. The prototype contrastive objective encourages different demonstrations of the same task to share a compact representation, while gated AdaLN allows the velocity network to reduce prototype influence when the support signal is unreliable. The Supporting Material reports the corresponding perturbation diagnostics: as support perturbation increases, gate activation decreases and trajectory error grows more slowly than in direct prototype concatenation. In this sense, the gate is not only a network component but also a robustness mechanism that prevents few-shot conditioning from becoming a noise amplifier.

\noindent\textbf{Efficiency versus raw in-context conditioning.}
The comparison with In-Context Flow highlights an important practical trade-off. Keeping raw support tokens online preserves detailed demonstration information, but the context length increases with the number and duration of demonstrations. This raises latency and memory usage, which is undesirable for closed-loop robot control. Pooled-Demo CFM has the opposite behavior: it is fast, but the single vector loses temporal and phase-level structure. TP-Flow occupies a more useful middle ground. The support set is encoded once into a small number of prototype tokens, and the online velocity network conditions on this fixed-size representation. Table~\ref{tab:efficiency} and Fig.~\ref{fig:inference-efficiency} show that this design keeps the policy within the 10 Hz control target while preserving much of the adaptation benefit of demonstration conditioning.

\noindent\textbf{When TP-Flow may fail.}
The current framework still has several limitations that suggest clear directions for future work. First, cross-embodiment evaluation on robots with different kinematics, grippers, cameras, and control frequencies remains an important next step beyond the present LEROBOT-ARM-SO101 study. Second, the method relies on the support set being broadly consistent with the query task. If the demonstrations contain the wrong goal, a missing phase, or a misleading recovery behavior, the prototype can still bias the generated trajectory in the wrong direction. Third, contact-rich tasks with deformable objects, heavy occlusion, or fine force regulation may require tactile or force feedback beyond the current visual-proprioceptive observation stream. In our failure observations, wrong grasp pose, contact instability, prototype mismatch, over-confident priors, and insufficient ODE integration remain important cases for future improvement.

\noindent\textbf{Implications for future VLA manipulation.}
The broader implication is that few-shot robot learning does not need to rely only on large-scale fine-tuning or ever-longer context windows. A more scalable direction is to transform demonstrations into compact, structured, and controllable task representations that can be reused by a generative action model. TP-Flow takes one step in this direction by connecting structured prototype encoding, adaptive flow priors, gated velocity modulation, and theoretical diagnostics. Future extensions could study active support-demonstration selection, cross-embodiment prototype alignment, tactile-aware prototype tokens, and integration with larger VLA backbones. These directions would make the prototype not only a task descriptor, but a bridge between high-level instruction understanding and low-level contact-aware execution.

\section{Conclusion}\label{sec:conclusion}

This paper presented TP-Flow, a task-prototype guided flow matching framework for few-shot vision-language robot manipulation. The key idea is to transform a small support set into structured task-prototype tokens and use these tokens to guide both the initial flow prior and the velocity field. Unlike ordinary conditional flow matching, TP-Flow does not treat demonstrations as a single pooled condition or as a long online context. Instead, it extracts phase-level execution structure through symmetric cross-attention, parameterizes a task-adaptive prior, and modulates the velocity network through gated adaptive normalization.

Experiments on the LEROBOT-ARM-SO101 platform demonstrated that this design improves few-shot generalization across held-out target tasks, novel objects, goal recombination, long-horizon execution, and noisy-support settings. The results showed that TP-Flow achieves stronger success rates than behavior-cloning, VLA-style, action-chunking, diffusion-policy, and non-prototype flow-matching baselines, while maintaining real-time closed-loop execution. Ablation studies further confirmed that structured prototypes, the reverse attention path, the adaptive prior, gated modulation, and prototype contrastive regularization each contribute to the final performance.

The Supporting Material provides the detailed theoretical analysis and diagnostic experiments behind these empirical gains. The diagnostics show that prototype distances align with action-distribution distances, the adaptive prior reduces transport cost and flow-matching loss variance, and gated modulation attenuates support-set perturbations while keeping measured trajectory deviations below the derived ODE bound. These findings suggest that task prototypes are not only useful conditioning features, but also measurable representations of task-level execution structure. Future work will extend TP-Flow to cross-embodiment transfer, tactile and force-aware manipulation, active support-demonstration selection, and more complex contact-rich tasks.

\noindent\textbf{Acknowledgements}
Not applicable.

\section*{Declarations}

\noindent\textbf{Conflict of interest}
The authors declare no conflict of interest.

\medskip
\noindent\textbf{Ethics approval and consent to participate}
Not applicable.

\medskip
\noindent\textbf{Consent for publication}
Not applicable.

\medskip
\noindent\textbf{CRediT authorship contribution statement}
Omitted for anonymous review.

\printbibliography

\end{document}